\documentclass[10pt]{article} 
\usepackage[preprint]{tmlr}

\usepackage{amsmath,amsfonts,bm}

\def\eqref#1{equation~\ref{#1}}

\def\1{\bm{1}}

\DeclareMathAlphabet{\mathsfit}{\encodingdefault}{\sfdefault}{m}{sl}
\SetMathAlphabet{\mathsfit}{bold}{\encodingdefault}{\sfdefault}{bx}{n}

\usepackage{hyperref}
\usepackage{url}
\usepackage{graphicx}%
\usepackage{multirow}%
\usepackage{amsmath,amssymb,amsfonts}%
\usepackage{mathtools}      
\usepackage{amsthm}%
\usepackage{mathrsfs}%
\usepackage[title]{appendix}%
\usepackage{xcolor}%
\usepackage{textcomp}%
\usepackage{manyfoot}%
\usepackage{booktabs}%
\usepackage{algorithm}%
\usepackage{algorithmicx}%
\usepackage{afterpage}
\usepackage{algpseudocode}%
\usepackage{listings}%
\usepackage{tabularx}
\usepackage{bookmark}
\usepackage{longtable}
\usepackage{anyfontsize}
\usepackage{xcolor}
\usepackage{siunitx}
\usepackage{subcaption}
\usepackage{xcolor}         
\usepackage{colortbl}
\usepackage{float}
\usepackage{wrapfig}
\newcolumntype{K}[1]{>{\centering\arraybackslash}p{#1}}
\definecolor{mygray}{gray}{0.8}
\definecolor{mygray2}{gray}{0.9}

\title{Benchmarking Stochastic Interpolants for \\ Modeling Physical Systems}

\author{\name Anthony Zhou \email ayz2@andrew.cmu.edu \\
      \addr Department of Mechanical Engineering\\
      Carnegie Mellon University
      \AND
      \name Alexander Wikner \email awikner@uchicago.edu \\
      \addr Department of Geophysical Sciences \\
      University of Chicago
      \AND
      \name Amaury Lancelin \email lancelin.amaury@lmd.ipsl.fr\\
    \addr  LMD/IPSL, CNRS, ENS, Universit\'e PSL \\ 
    \'Ecole Polytechnique, Institut Polytechnique de Paris, Sorbonne Universit\'e \\ 
    R\'eseau de Transport d’Electricit\'e (RTE) 
    \AND
      \name Pedram Hassanzadeh \email pedramh@uchicago.edu \\
      \addr Department of Geophysical Sciences \\ 
      Committee on Computational and Applied Mathematics \\ 
      University of Chicago
      \AND
      \name Amir Barati Farimani \email barati@cmu.edu \\
      \addr Department of Mechanical Engineering\\
      Carnegie Mellon University
      }

\def\month{MM}  
\def\year{YYYY} 
\def\openreview{\url{https://openreview.net/forum?id=XXXX}} 

\begin{document}

\maketitle

\begin{abstract}
Generative models have recently emerged as powerful surrogates for physical systems, demonstrating increased accuracy, stability, and/or statistical fidelity. However, most approaches rely on iteratively denoising a Gaussian, a choice that may not be the most effective for autoregressive prediction tasks, where the current and future states are often closely related. We therefore benchmark generative models based on stochastic interpolants, which can learn maps between arbitrary densities, against standard Gaussian-denoising frameworks across a range of fluid and climate systems. Following prior work, we consider a transport map for dynamical systems that learns an interpolant between a point mass at the current state and a conditional distribution of future states. We find that stochastic interpolants can match or outperform Gaussian-denoising baselines in pointwise accuracy, spectral fidelity, and probabilistic skill. Furthermore, we identify a tradeoff where deterministic sampling can improve pointwise accuracy but smooth predictions, whereas stochastic sampling recovers spectral content and calibration but requires more function evaluations, as well as ensemble averaging to recover pointwise accuracy. As a whole, these results establish stochastic interpolants as a competitive baseline for generative physical emulation, and also indicate that the training loss, sampler, and transport map should be chosen with the physical system and target metric in mind.
\end{abstract}

\section{Introduction}\label{sec:introduction}

The field of neural PDE surrogates has grown immensely since its initial foundations \citep{li2021fourierneuraloperatorparametric, Lu_2021}, with a wide range of architectures now being applied across many scientific and engineering applications \citep{ohana2025welllargescalecollectiondiverse}. Within this landscape, generative models have emerged as a particularly promising class of methods for modeling physical systems, achieving state-of-the-art results in modeling phenomena such as turbulence \citep{molinaro2025generativeaifastaccurate} and weather \citep{price2024gencastdiffusionbasedensembleforecasting}. These results are supported by many empirical findings, where diffusion models have demonstrated better accuracy and stability \citep{lippe2023pderefinerachievingaccuratelong}, capable of resolving finer details than deterministic baselines \citep{oommen2025integratingneuraloperatorsdiffusion}. In addition, diffusion models can be more effective at capturing the underlying statistics of physical systems, such as in turbulence \citep{Lienen2023, molinaro2025generativeaifastaccurate} or in weather forecasting and climate prediction \citep{price2024gencastdiffusionbasedensembleforecasting, cachay2023dyffusiondynamicsinformeddiffusionmodel}. Overall, these capabilities are supported by studies that benchmark diffusion models in PDE systems \citep{kohl2024benchmarkingautoregressiveconditionaldiffusion, rozet2025lostlatentspaceempirical}, as well as by continued research that improves their accuracy or inference speed \citep{bastek2025physicsinformeddiffusionmodels, shehata2025improved}. Moreover, these models continue to benefit from larger advances in the generative modeling community, such as flow matching or improved samplers \citep{liu2022flowstraightfastlearning, Lu_2025}. 

While promising, a key feature of prior generative models in PDE or climate modeling is the assumption of a Gaussian prior/source distribution. This is a logical choice for unconditional generation, where we have no prior knowledge about the source distribution and require it to be easily sampled. However, recent work has challenged this assumption for tasks where the source and target distributions are related, such as in image-to-image translation or super-resolution. Ordinarily, these tasks are framed as sampling from a Gaussian and evolving a reverse process conditioned on the source distribution, however, methods such as diffusion bridges or stochastic interpolants seek to directly learn a stochastic process between the source and target distributions \citep{zhou2023denoisingdiffusionbridgemodels, albergo2023buildingnormalizingflowsstochastic}. Directly evolving samples drawn from the source distribution (e.g., blurry images, masked images) to samples from the target distribution (e.g., sharp images, in-painted images) can require fewer sampling steps and produce higher quality samples \citep{albergo2024stochasticinterpolantsdatadependentcouplings, zheng2025diffusion}. In these cases, it is believed that transporting Gaussian noise to a conditional distribution is more complex than directly mapping the source to the target distribution. 

This observation motivates stochastic interpolants as a generative framework for physical systems. Predicting PDE or climate systems is usually framed as an autoregressive task, where future states are predicted from a current state. Because successive states are often tightly coupled, transporting Gaussian noise while conditioning on the current state may inefficient, whereas an interpolant can exploit this proximity by mapping directly between the current and future distributions. To test whether this helps in practice, we benchmark stochastic interpolants against diffusion and flow matching frameworks on three systems: Kolmogorov flow, Rayleigh–Bénard convection, and medium-range weather forecasting and long-term climate emulation with PlaSim. To isolate the effect of the generative framework, all models share the same latent space, diffusion-transformer backbone, and training budget, and are evaluated at standardized sampling budgets.  Our experiments show that stochastic interpolants can be a strong baseline, and attains the lowest single-sample error on both fluid benchmarks. Additionally, it attains the best CRPS and near-ideal spread–skill ratios across all three systems, and the lowest 10-year climatological biases in climate emulation. More broadly, we also find that generative models can trade off deterministic accuracy, spectral consistency, and probabilistic calibration depending on their sampling. Together, these results offer practical guidance for choosing and sampling generative surrogates, and suggest that there are future opportunities to design transport maps tailored to the physical system and task of interest.

\section{Related Work \label{sec:related}}
\paragraph{Generative Models}
Modern generative models can be understood as the dynamical transport of one probability density into another. Diffusion
models learn to reverse a forward noising process that gradually corrupts data
into Gaussian noise, formulated either as denoising diffusion
\citep{ho2020denoisingdiffusionprobabilisticmodels} or through the lens of
score matching and stochastic differential equations
\citep{song2021scorebasedgenerativemodelingstochastic}. Subsequent work
improved the sampling process, introducing deterministic and accelerated
samplers \citep{song2022denoisingdiffusionimplicitmodels} and a principled
study of the design space governing noise schedules, preconditioning, and
discretization \citep{karras2022elucidatingdesignspacediffusionbased}. Flow
matching offers another alternative, regressing directly onto the velocity field of
a probability path to bridge exactly between two densities in finite time
\citep{lipman2023flowmatchinggenerativemodeling}, with rectified flows further
straightening transport trajectories to enable few-step sampling
\citep{liu2022flowstraightfastlearning}. A common thread is that these methods
transport a fixed Gaussian prior to a target density in order to sample from it.

A second line of work relaxes the requirement of a Gaussian prior, allowing
transport between two arbitrary densities. Such formulations arise under several
related frameworks, including Schr\"odinger Bridges
\citep{debortoli2023diffusionschrodingerbridgeapplications}, Diffusion Bridges
\citep{su2023dualdiffusionimplicitbridges, zhou2023denoisingdiffusionbridgemodels},
Optimal Transport \citep{tong2024improvinggeneralizingflowbasedgenerative}, and
Stochastic Interpolants \citep{Albergo2023}.
The stochastic interpolant framework in particular unifies flow- and
diffusion-based models by constructing an explicit bridge between any two
densities in finite time and derives training objectives and sampling algorithms to both learn this interpolant and allow deterministic flows or stochastic samplers. Viewed through this lens,
widely used methods such as DDPM and EDM can be seen as particular
instantiations of a more general principle of dynamical transport between
distributions. Together, these theoretical and practical foundations have greatly expanded the capabilities of generative models and enabled a wide range of applications. In particular, generative models have seen widespread adoption in predicting and modeling important physical phenomena.

\paragraph{Deterministic Neural Surrogates}
Deep learning surrogates for PDE and climate systems are a diverse and rapidly
growing field, where surrogate models generally learn a deterministic map from the current
system state to a future state. Many of these methods can be cast as neural operators, which learn mappings between
infinite-dimensional function spaces and therefore generalize across discretizations \citep{li2021fourierneuraloperatorparametric,Lu_2021, JMLR:v24:21-1524}.
Building on this paradigm, a wide range of architectures have been proposed,
including convolutional and spectral models \citep{pathak2022fourcastnet, raonic2023convolutional}, transformer- and attention-based
operators
\citep{ li2023transformerpartialdifferentialequations, bi2022panguweather3dhighresolutionmodel, li2024cafaglobalweatherforecasting}, the use of 
large-scale pretraining and foundation models
\citep{zhou2024maskedautoencoderspdelearners, mccabe2024multiplephysicspretrainingphysical, zhou2024strategiespretrainingneuraloperators, nguyen2023climaxfoundationmodelweather},
and models that use physics-based priors or inductive biases
\citep{verma2024climodeclimateweatherforecasting, zhou2025hamiltonianneuralpdesolvers}.

For time-dependent systems, these surrogates are typically queried autoregressively, applying a learned single-step operator repeatedly to roll out a trajectory. While effective at minimizing pointwise error over short horizons, there are several well-studied failure modes. Models are generally trained on ground-truth inputs but use their own predictions at inference time; therefore, the input distribution
gradually drifts from the training distribution and causes errors to compound
over long rollouts \citep{lippe2023pderefinerachievingaccuratelong, LIST2025117441}. This is compounded by the fact that models trained with mean-squared-error losses are subject to a spectral bias that prioritizes low-frequency modes, yielding
predictions that regress toward the conditional mean and progressively lose high-frequency content. This issue is especially prevalent in chaotic and turbulent dynamics, resulting in model predictions that smooth over time or autoregressive instability, as errors in the high-frequency band that eventually transport to the lower frequencies \citep{mccabe2023stabilityautoregressiveneuraloperators}. Lastly, a deterministic surrogate produces only a single trajectory, therefore, it cannot represent the uncertainty inherent to chaotic systems or provide calibrated probabilistic forecasts. These limitations have motivated the use of generative models in certain physical systems, 
which instead learn to sample from a distribution over future states.

\paragraph{Generative Surrogates for Physical Systems}
Physical systems can be seen as probabilistic in many practical settings. A lack of spatial resolution, the need to ignore or approximate sub-grid effects, partial observation, or chaotic systems  all contribute to underdetermined systems where  the map from the current state
to the future is multimodal and is poorly approximated by a point estimate.
Generative surrogates address this directly by learning the full conditional
distribution of future states given the current state, $p(u_{t+1} \mid u_t)$,
and sampling from it rather than regressing onto its mean. Because individual
samples are drawn from the learned data distribution, they retain sharp
small-scale structure and physically consistent energy spectra instead of the
over-smoothed fields that result from minimizing a mean-squared-error loss. This makes generative models compelling for turbulent and multiscale systems, where spectral and statistical fidelity are important, and a growing body of work has shown that they can match or
exceed deterministic surrogates on both accuracy and distributional metrics
\citep{oommen2025integratingneuraloperatorsdiffusion, Shu2023, cachay2023dyffusiondynamicsinformeddiffusionmodel}. Generative models have also been shown to remain more stable than their deterministic counterparts during autoregressive rollout \citep{kohl2024benchmarkingautoregressiveconditionaldiffusion, pedersen2025thermalizerstableautoregressiveneural, liu2026infinitelylongneuralsimulations}; while the precise mechanism is unknown, it is hypothesized that the injection of noise at each timestep continuously perturbs the state and reintroduces high-frequency modes into the rollout that may otherwise decay over time. 

Many prior works have applied diffusion models to predict
PDE systems
\citep{yang2023denoisingdiffusionmodelfluid, huang2024diffusionpdegenerativepdesolvingpartial, Shu2023, cachay2023dyffusiondynamicsinformeddiffusionmodel, du2024confildconditionalneuralfield, serrano2024aromapreservingspatialstructure, molinaro2025generativeaifastaccurate, Gao2024_comm},
typically by conditioning the generative process on the current state and sampling successive states autoregressively. Beyond
direct prediction, these models have been adapted to the structure of physical
problems in several ways, such as incorporating physics-informed losses to enforce
governing equations \citep{bastek2025physicsinformeddiffusionmodels}, performing diffusion in wavelet or spectral space \citep{hu2025waveletdiffusionneuraloperator, gallon2026physicsinformeddiffusionmodelsspectral}, operating
on irregular meshes to handle complex geometries
\citep{lino2025learningdistributionscomplexfluid}, and allowing text-conditioned
generation of physical fields \citep{zhou2025text2pdelatentdiffusionmodels}. Progress in the broader field of generative models, such as latent diffusion \citep{rombach2022highresolutionimagesynthesislatent, rozet2025lostlatentspaceempirical}, accelerated samplers \citep{shehata2025improved}, or sampling guidance \citep{yao2026guideddiffusionsamplingfunction}, have also improved performance in physics-based problems. Recent progress in flow matching has also been reflected in PDE emulation,
where learning a velocity field along a straightened probability path can yield
additional speed or accuracy benefits 
\citep{li2025generativelatentneuralpde,baldan2025flowmatchingmeetspdes, utkarsh2025physicsconstrainedflowmatchingsampling,armegioiu2025rectifiedflowsfastmultiscale,shi2024universalfunctionalregressionneural}.

Generative approaches are also prevalent and successful in weather forecasting, where
the inherently probabilistic, chaotic nature of the atmosphere makes ensemble
generation a natural fit and established diffusion and flow matching models as
standard tools
\citep{price2024gencastdiffusionbasedensembleforecasting,zhuang2025ladcastlatentdiffusionmodel, cachay2024probabilisticemulationglobalclimate, couairon2024archesweatherarchesweathergendeterministic, cachay2025elucidatedrollingdiffusionmodels}. Indeed, current state-of-the-art weather forecasting models incorporate probabilistic or generative components \citep{alet2025skillfuljointprobabilisticweather, lang2024aifscrpsensembleforecastingusing} and are being deployed to complement numerical weather prediction in operational forecasts. There are also several practical tasks in weather forecasting that are ill-posed or inverse problems, where many states are
consistent with the available information. In these tasks, such as downscaling or data assimilation, generative models have also shown strong performance 
\citep{mardani2024residualcorrectivediffusionmodeling, gao2023prediffprecipitationnowcastinglatent, andry2025appabendingweatherdynamics, aich2024conditionaldiffusionmodelsdownscaling}.

These prior works highlight the unique capabilities of generative models and identify applications and problems for which they are best suited. For chaotic or underdetermined systems, generative models can be a natural and effective choice. These systems exhibit forecast uncertainty that grows with lead time, and generative models can produce calibrated ensembles
and uncertainty estimates that deterministic models cannot. Furthermore, for multiscale or turbulent systems, generative models can reproduce small-scale structure and energy spectra by sampling from the conditional distribution and avoiding the spectral collapse of mean-regression. Lastly, there are emergent phenomena such as increased autoregressive stability that are beneficial. However, the principal cost of these benefits is the increased inference time and training complexity. Generating a sample requires integrating a learned dynamical process over many steps, which can be orders of magnitude slower than a single deterministic forward pass. Furthermore, a wide variety of training losses and parameterizations have been proposed due to the complexity of learning a probabilistic map, as opposed to minimizing a more straightforward mean-squared regression objective. These practical limitations serve as the motivation for this work, where we seek to investigate if a principled choice of transport map, rather than a fixed Gaussian
prior, can improve inference speed and overall performance. 

\paragraph{Stochastic Interpolants for Physical Systems}
While diffusion and flow matching models based on Gaussian denoising are now
well established and validated, interpolant-based approaches that map between
arbitrary distributions remain comparatively rare in physical systems. Several
works have applied them to super-resolution and data assimilation in PDEs and
climate
\citep{bischoff2023unpaireddownscalingfluidflows, Rout2025, schiødt2025generativesuperresolutionturbulentflows, chen2025flowdasstochasticinterpolantbasedframework},
motivated by the resemblance of these problems to the denoising and inpainting
tasks where stochastic interpolants have proven effective in computer vision.
Most directly related to our work,
\citet{chen2024probabilisticforecastingstochasticinterpolants} apply interpolants
to forecasting dynamical systems, learning a transport map from a point mass at
the current state to the conditional distribution of future states and developing much of the supporting theory. Complementary theoretical advances have since refined the design of the interpolant itself, for example through
scale-adaptive flows for multiscale scientific data
\citep{chen2026scaleadaptivegenerativeflowsmultiscale} and principled choices of
interpolation schedule \citep{chen2026lipschitzguideddesigninterpolationschedules}.

Whereas these prior works establish the theoretical foundations of generative models and stochastic interpolants for dynamical systems, our contribution is primarily empirical. Many theoretical frameworks for generative modeling now exist, but it remains unclear which are best suited to physical systems or what each one offers. Adding to this, physical emulation exhibits remarkable fragmentation: different systems exhibit different dynamics, variables, and spatial or temporal scales, and even within the same system, practitioners may prioritize very different objectives. In fluid dynamics, for instance, the Reynolds number can fundamentally reshape the relevant length and time scales. Even when the governing system is fixed, such as the Earth's climate, a weather forecaster may seek to minimize medium-range RMSE, whereas a climate researcher may focus on reducing long-term climatological biases. This diversity suggests that generative models may need to tailor their training or sampling to the demands of each problem, which in turn presents an opportunity to develop empirically grounded heuristics and best practices for these settings. We therefore conduct a comprehensive benchmark of stochastic interpolants for physical emulation, with the goal of understanding when they work and how these benefits occur. Our central finding is that principled choices for the transport map and sampling algorithm yield meaningful performance gains over standard generative frameworks from image or video modeling. Across fluid and climate benchmarks, our experiments demonstrate that stochastic interpolants can outperform standard choices and offer helpful inference-time tradeoffs. In doing so, we establish stochastic interpolants as a competitive baseline for physical emulation and offer broader insight into the behavior of generative surrogates in this setting.
\section{Methods \label{sec:methods}}
\subsection{Stochastic Interpolants}
\label{sec:SI}
\paragraph{Overview} We consider a class of generative models that admit arbitrary source and target distributions. These models have various names and instantiations, however, for simplicity, we consider stochastic interpolants since many of these models can be unified under this framework \citep{Albergo2023, zhang2025exploringdesignspacediffusion}. Given samples \(x_0\) from a source distribution \(\rho_0\) and samples \(x_1\) from a target distribution \(\rho_1\), a stochastic interpolant \(x_t\) is defined as:
\begin{equation}
    x_t = \alpha(t)x_0 + \beta(t)x_1 + \gamma(t)z, \qquad t\in [0, 1]
\end{equation}
The mixing coefficients \(\alpha(t), \beta(t), \sigma(t)\) must satisfy the boundary conditions \(\alpha(0) = \beta(1) = 1\),  \(\alpha(1) = \beta(0) = 0\), and \(\gamma(0) = \gamma(1) = 0\) to recover the endpoints of the trajectory \(x_0, x_1\). Furthermore \(z\) is generally sampled from a standard Gaussian \(z \sim N(0, I)\), and serves to mix the intermediate distributions \(p_t(x)\). There are a few useful observations to make. Firstly, the stochastic interpolant produces samples \(x_0 \sim \rho_0\) at \(t=0\) and \(x_1 \sim \rho_1\) at \(t=1\) by construction. Furthermore, the interpolant \(x_t\) maps between the densities \(\rho_0,\rho_1\) exactly and in finite time. Lastly, this stochastic interpolant can be realized by an SDE or ODE, which can produce samples \(x_t\) from intermediate distributions \(p_t(x)\) at any time \(t\in[0, 1]\). 

\paragraph{Implementation} 
The overall framework of stochastic interpolants is designed to be broad, with the goal of unifying different generative modeling frameworks. However, for its practical deployment in physical systems, we must make specific choices of the interpolant \(x_t\), as well as specify training and sampling algorithms. The key inductive bias that is leveraged is to consider a source distribution of a point mass centered at the current state \(x_0 \sim \delta (u_t)\) and a target distribution of future states conditioned on the current state \(x_1 \sim p(u_{t+1} | u_t)\), both supported on \(\mathbb{R}^d\) for \(d\)-dimensional states. The hypothesis is that this reduces the transport cost compared to a Gaussian prior, which can simplify both the learning the transport map and sampling from a trained model. This choice of a Dirac source distribution induces the interpolant \citep{chen2024probabilisticforecastingstochasticinterpolants}:
\begin{equation}
    \label{eq:si_eqn}
    x_t = \alpha(t)x_0 + \beta(t)x_1 + \sigma(t)W_t, \qquad t\in [0, 1]
\end{equation}
where \(W_t\) is a Wiener process and is realized with \(W_t \stackrel{d}{=} \sqrt{t}z\) by Brownian increments generated from \(z\sim N(0, \mathbf{I})\), and \(\sigma(t)\) is a noise coefficient that vanishes at \(t=1\). We further specify linear choices of coefficients \(\alpha(t) = 1-t\), \(\beta_t = t\), and \(\sigma(t) = 1-t\) which gives:
\begin{equation}
    x_t = (1-t)x_0 + tx_1 + (1-t)W_t, \qquad t\in [0, 1]
\end{equation}
To realize the interpolant \(x_t\), we parameterize a drift \(b_\theta\), which is learned by minimizing the loss:
\begin{equation}
    \mathcal{L}_b[b_\theta] = \frac{1}{N}\sum_{n=1}^N
    ||b_\theta(t,x^n_{t}, x_0^n) - R(t, x^n_{1}, x_0^n))||^2
\end{equation}
where 
\begin{equation}
    R(t, x^n_{1}, x_0^n) = \dot \alpha (t)x_0^n + \dot \beta(t) x_1^n + \dot{\sigma}(t)W_t 
\end{equation}
The timestep \(t \in [0,1]\) is uniformly sampled and \(x^n_{t}\) is a given data sample \(n\) at time \(t\) along the stochastic interpolant. Additionally, to map to a conditional distribution, the condition \(x_0^n\) is provided as input to the estimated drift. Under linear choices of mixing coefficients, the training target reduces to:
\begin{equation}
    \mathcal{L}_b[b_\theta] = \frac{1}{N}\sum_{n=1}^N
    |b_\theta(t,x^n_{t}, x_0^n) - (x_1^n - x_0^n - W_t)|^2
\end{equation}
which can be computed by sampling \(x_0^n = u_t\), \(x_1^n = u_{t+1}\) from a dataset (where \(t\) denotes physical time) and \(W_t \stackrel{d}{=} \sqrt{t}z\) from \(z \sim N(0, I)\). Intuitively, \(b_\theta\) aims to estimate the time derivative of the stochastic interpolant. This is useful during inference, where samples \(x_0\) are integrated using conditional drift estimates \(b_\theta\) according to the probability flow SDE, starting from \(t=0\) to \(t=1\):
\begin{equation}
    \label{eqn:sampling_sde}
    dx_t^{SDE} = b_\theta(t, x_t, x_0)dt + \sigma(t)dW_t
\end{equation}
At \(t=1\), this SDE produces samples from the conditional distribution \(p(x_1|x_0)\). To realize this SDE, we use a numerical integrator, specifically the Euler-Maruyama method:
\begin{equation}
    \quad x_{t+1} = x_t + \Delta t * b_\theta(t, x_t, x_0) + (1-t)\sqrt{\Delta t} * z
\end{equation}
where a linear choice of \(\sigma(t)\) was substituted and timesteps \(\Delta t\) are uniformly discretized on \([0, 1]\). We can observe that at \(t=0\), the diffusion term \(\sigma(t)dW_t\) is maximized to immediately spread the probability mass of the Dirac \(x_0 \sim \delta(u_t)\). As \(t \rightarrow 1\), the diffusion term vanishes to recover clean samples from the conditional distribution \(x_1 \sim p(u_{t+1}|u_t)\). 

\paragraph{Variable Diffusion}
Interestingly, this framework admits a family of SDEs with variable noise coefficients \(\hat \sigma(t)\) by constructing a modified drift \(\hat b_\theta (t, x_t, x_0)\). We refer to \citet{chen2024probabilisticforecastingstochasticinterpolants} for details, but under regularity conditions, SDEs with variable diffusion can be sampled without retraining \(b_\theta (t, x_t, x_0)\). For future states with more or less uncertainty, this can modify the stochasticity of the sampler at test time. 

In our case, we allow the noise coefficient to be scaled by a constant \(\hat \sigma (t) = \sigma (1-t)\) during sampling. While this requires constructing a correction to the drift \(\hat b_\theta (t, x_t, x_0)\), we find that empirically, using the same drift without modification yields similar results and simplifies the sampling algorithm. One interesting case is when \(\sigma(t) \rightarrow 0\) for all \(t\). In this case, the SDE reduces to an ODE which generally allows for faster sampling by truncating the reverse process. However, this sampler degenerates the interpolant to predict the deterministic mean of the conditional distribution. For most generative models (e.g., in image or text generation) this is detrimental due to the multimodal nature of the target distribution, which would generate a blurry, average sample. However, certain physics applications have strongly unimodal target distributions, where future states are almost deterministic when conditioned on the present. This can happen if the temporal dynamics are predictable or the lead time between states is small. From this point of view, we can see that adjusting \(\sigma(t)\) in Equation \ref{eqn:sampling_sde} forms a tradeoff: at small \(\sigma(t)\) we produce a deterministic neural surrogate, and at large \(\sigma(t)\) forward diffusion dominates the SDE. 

\subsection{Generative Models}
\paragraph{Overview}
In addition to the constructed stochastic interpolant, we also train a variety of commonly used diffusion and flow matching models. An overview of the relevant training and sampling procedures is given in Table \ref{tab:frameworks}. Denoising Diffusion Probabilistic Models (DDPMs) \citep{ho2020denoisingdiffusionprobabilisticmodels} define a fixed Markovian forward process that gradually corrupts a data sample \(x_0\) into Gaussian noise over \(T\) steps and trains a network to approximate the reverse transitions, typically by predicting the noise added at each step. While this yields high-quality outputs, it requires a large number of function evaluations (NFEs), which is a limitation for autoregressive physics surrogates. Denoising Diffusion Implicit Models (DDIMs) \citep{song2022denoisingdiffusionimplicitmodels} retain the DDPM training objective but uses a deterministic reverse process that can be sampled in fewer steps, which trades sample quality for inference speed. Elucidated Diffusion Models (EDMs) \citep{karras2022elucidatingdesignspacediffusionbased} reframe diffusion within a continuous-time framework that disentangles the the noise schedule, network preconditioning, loss weighting, and sampling; optimizing these results in achieving high fidelity at comparatively few NFEs. Truncated Sampling Models (TSMs) \citep{shehata2025improved} target the inference cost of DDPMs directly by truncating the low-noise regime of the reverse process. Sampling is carried out only down to an intermediate step  after which the clean state is estimated via Tweedie's formula. For unimodal, conditional distributions in autoregressive prediction tasks, this enables single- and few-step sampling while preserving or even improving accuracy relative to full DDPM sampling. Lastly, flow matching (FM) \citep{liu2022flowstraightfastlearning, lipman2023flowmatchinggenerativemodeling} seeks to learn a continuous-time velocity field that transports samples from a standard Gaussian to the target distribution. Rather than relying on  stochastic denoising in diffusion, flow matching trains a network to directly regress a vector field and generates samples by integrating the resulting ODE from noise to data. When the probability path is constructed to linearly interpolate between the two distributions, the induced trajectories are nearly straight which enables accurate generation with comparatively few NFEs. 

One remark is that the general framework of stochastic interpolants can encompass these standard generative models, however, for clarity we refer to the specific interpolant in Equation \ref{eq:si_eqn} as the SI model going forward. 

\begin{table}[t!]
\centering
\small 
\def\arraystretch{1.5}
\begin{tabular}{lll}
\toprule
 & \textbf{Training Objective} & \textbf{Sampling Procedure} \\
\midrule
\textbf{DDPM} &
\(
\mathcal{L} = 
\left\|
\epsilon - \epsilon_\theta\!\left(x_t,c,  t
\right)
\right\|^2
\)
& 
\(
x_{t-1} =
\frac{1}{\sqrt{\alpha_t}}\!\left(
x_t - \frac{1-\alpha_t}{\sqrt{1-\bar{\alpha}_t}}\,\epsilon_\theta(x_t,c,t)
\right) + \sigma_tz
\)
\\
\midrule
\textbf{DDIM} &
\(
\mathcal{L} = 
\left\|
\epsilon - \epsilon_\theta\!\left(x_t,c, t
\right)
\right\|^2
\) &
\(
\begin{aligned}
    x_{t-1} &= \sqrt{\alpha_{t-1}}\left( \frac{x_t - \sqrt{1-\alpha_t}\epsilon_\theta(x_t,c, t)}{\sqrt{\alpha_t}}\right) \\&+ \sqrt{1-\alpha_{t-1} - \sigma_t^2} \cdot \epsilon_\theta(x_t, c,t) + \sigma_t z
\end{aligned}
\)
\\
\midrule
\textbf{EDM} &
\(
\mathcal{L} = 
\left\|
D_\theta(x+z,c,\sigma) - x
\right\|^2
\)
& 
\(
x_{t+\Delta t} = x_t +
\frac{\dot{\sigma}(t)}{\sigma(t)}
\left(x - D_\theta(x,c,\sigma(t))\right)\Delta t \:
\) \footnotesize(Euler)
 \\
\midrule
\textbf{TSM} &
\(
\mathcal{L} =
\left\|
\epsilon - \epsilon_\theta\!\left(x_t,c, t
\right)
\right\|^2
\)
& 
\(
x_{0} =
(x_t - \sqrt{1-\bar{\alpha}_t}\epsilon_\theta(x_t, c,t))/\sqrt{\bar{\alpha}_t} \:
\) \footnotesize(1-step)\\ 
\midrule
\textbf{FM}&
\(
\mathcal{L} = \left\|
(x - z) - v_\theta\!\left(x_t,c, t\right)
\right\|^2 
\)
& 
\(
x_{t+\Delta t} = x_t + v_\theta(x_t,c, t)\Delta t, \: x_0\sim\mathcal{N}(0, I) \:
\) \footnotesize (Euler) \\
\midrule
\textbf{SI}&
\(
\mathcal{L} = \left\|
((x_1 - x_0 - W_t) - b_\theta \left(x_t, c, t\right))
\right\|^2 
\)
& 
\(
x_{t+\Delta t} = x_t + b_\theta(x_t,c,t)\Delta t + \sigma(t)\sqrt{\Delta t} z, \:
\) \footnotesize(Euler-Maruyama) \\
\bottomrule
\end{tabular}
\caption{Training and sampling for conditional diffusion, flow matching, and stochastic interpolant frameworks. \(x_t\) denotes the noised state, \(c\) denotes any conditioning information, and \(t\) denotes the diffusion or interpolant timestep. The noised state \(x_t\) is defined differently for diffusion, flow matching, and stochastic interpolants, and different samplers can be used for EDM and SI.}
\label{tab:frameworks}
\end{table}

\paragraph{Implementation}

DDPM, DDIM, and TSM models use some variant of the stochastic process:
\begin{equation}
    x_t = \sqrt{\bar \alpha_t} x_0 + \sqrt{1-\bar \alpha_t} z, \quad z\sim \mathcal{N}(0, \mathbf{I})
\end{equation}
which noises data \(x_0\) over some time \(t\in [1, T]\), where the noise schedule \(\beta_t = 1-\alpha_t\) is defined linearly with \(\beta_{start} = 0.0001\) and \(\beta_{end} = 0.02\). We found that scaling the noise schedule by \(S=400\) in \(S *\beta_{start}\), \(S *\beta_{end}\) to match the magnitude of the data was beneficial, following \citet{shehata2025improved}. In EDM, this stochastic process is reparameterized as:
\begin{equation}
    x_t = x_0 + \sigma_tz
\end{equation}
for a time-varying noise variance \(\sigma_t\). The design choices for EDM are already optimized, therefore, we did not find any substantial benefits by adjusting them. Ablations for DDPM/EDM hyperparameters can be found in Appendix \ref{app:ablation}. For flow matching, we adopt the commonly used linear schedule, which simplifies many hyperparameter choices and results in the interpolant:
\begin{equation}
    x_t = (1-t) x_0 + t z
\end{equation}
where the source distribution is taken to be a Gaussian. This can be seen as a counterpart to the SI model, where the source distribution is a Gaussian rather than a Dirac distribution centered at the current state. 

During inference, each generative model proposes a different sampler, which is generally fixed. However, the training and sampling for EDM/SI models is decoupled, which allows them to be sampled deterministically (Det.) or stochastically (Sto.). For EDM, this involves using the Euler method to integrate the probability flow ODE (Det.), or adding stochastic churn (Sto.). For SI models, this involves adjusting the noise coefficient during sampling, either using \(\sigma=0\) and sampling with the Euler method (Det.) or using \(\sigma=1\) and sampling with the Euler-Maruyama method (Sto.). Additional details on generative models are in Appendix \ref{app:generative}. 

\subsection{Models}
Within each generative model, the score, velocity, or denoiser is parameterized by a neural network. To ensure a consistent comparison, we standardize our model architectures and training procedures. We describe our approach here, but give additional details in Appendix \ref{app:arch}

\paragraph{Autoencoders} We choose to use latent diffusion \citep{rombach2022highresolutionimagesynthesislatent} to reduce the training and inference cost; however, in PDE problems it can also stabilize rollouts and improve performance \citep{rozet2025lostlatentspaceempirical, li2025latentneuralpdesolver}. For PDE tasks, we use a Deep Compression Autoencoder (DCAE) \citep{chen2025deepcompressionautoencoderefficient}, which is based on 2D convolution and residual pooling/unpooling layers. Following \citet{rozet2025lostlatentspaceempirical}, we apply a saturation function to latent vectors \(z = \frac{z}{\sqrt{1+z^2/b^2}}\), where \(b=5\), to avoid arbitrary variance. For climate tasks, we find that KL regularization is more beneficial, potentially due to the need for long-term consistency in climate emulators. For PDE problems we use a compression ratio of \(64\times\) and for climate problems we use a compression ratio of \(32\times\). 

\paragraph{Architectures} We use a diffusion transformer (DiT) \citep{peebles2023scalablediffusionmodelstransformers} as the backbone for all latent-space generative models. Within a given task, the architecture and model size is kept constant across all generative models. To condition on the diffusion or interpolant timestep, adaptive layer normalization is used. Additionally, the current state of the system is concatenated to the noisy, intermediate states to enable conditional generation. Moreover, as a result of the compressed latent space, the backbone does not need patchification or sparse attention. Although the work focuses on generative models, we also include a deterministic emulator as a baseline to understand the difficulty of tasks. For PDE tasks, we consider FNO \citep{li2021fourierneuraloperatorparametric} and for climate tasks, we consider SFNO \citep{bonev2023sphericalfourierneuraloperators}. All models are trained with an exponential moving average (EMA) of the weights and the Muon optimizer \citep{jordan2024muon}. 

\subsection{Datasets}
\paragraph{Kolmogorov Flow} Kolmogorov Flow (KF) is described by the 2D Navier-Stokes equations driven by unidirectional periodic forcing. In vorticity form, Kolmogorov Flow can be described by the PDE \citep{koehler2024apebenchbenchmarkautoregressiveneural}:
\begin{equation}
    \frac{\partial \omega }{\partial t} = -b\left( \begin{bmatrix}
           1 \\
           -1 \\
         \end{bmatrix} \odot \nabla (\Delta^{-1}\omega)
         \right) \cdot \nabla \omega + \nu \nabla \cdot \nabla \omega + \lambda \omega  - k \cos (k\frac{2\pi}{L}y)
\end{equation}
The leftmost term represents vorticity convection, controlled by the coefficient \(b\). Diffusion is controlled by the viscosity \(\nu\) and a drag term \(\lambda \omega \) is introduced. Lastly, a sinusoidal forcing term is introduced, controlled by the magnitude \(k\). 

Although common, this is a fairly challenging task and most PDE surrogates are unstable when rolled out to the training horizon \citep{lippe2023pderefinerachievingaccuratelong, Zhou_2025}. Data is generated from APEBench \citep{koehler2024apebenchbenchmarkautoregressiveneural} at a resolution of \(160\times160\) on a domain \((x, y) = [-10, 10]^2\). The simulation is saved at a resolution of \(\Delta t=0.2s\) for 100 timesteps, resulting in a rollout from \(t=0\) to \(t=20\) seconds. Initial conditions are uniformly sampled from a random truncated Fourier series with 5 modes, and the viscosity \(\nu\) is set to \(10^{-2}\) to simulate a Reynolds number of \(\sim 10^2\).

\paragraph{Rayleigh-Bénard Convection} Rayleigh-Bénard Convection (RBC) is a phenomenon that describes the mixing of horizontal layers of fluid driven by a temperature gradient. The equations are governed by a buoyancy and Navier-Stokes equation \citep{ohana2025welllargescalecollectiondiverse}:
\begin{align}
    \frac{\partial b}{\partial t} - \kappa \Delta b &= -u \cdot \nabla b \\ 
    \frac{\partial u}{\partial t} - \nu u + \nabla p - b\mathbf{e}_z &= -u\cdot \nabla u
\end{align}
The thermal diffusivity \(\kappa\) and viscosity \(\nu\) are determined by the Rayleigh and Prandtl numbers:
\begin{equation}
    \kappa = (\text{Rayleigh}\times \text{{Prandtl}})^{-\frac{1}{2}}, \quad \nu = \left(\frac{\text{Rayleigh}}{\text{Prandtl}}\right)^{-\frac{1}{2}}
\end{equation}
Rayleigh and Prandtl numbers are varied throughout the training and validation data, where \(\text{Rayleigh} \in \{1e6, 1e7, 1e8, 1e9, 1e10\}\) and \(\text{Prandtl} \in \{0.1, 0.2, 0.5, 1,2, 5, 10\}\). Furthermore, initial conditions for the buoyancy are generated by \(b(t=0)=(Ly-y)\times \delta b_0 +y(Ly-y)\times \epsilon\), where \(\delta b_0\) is sampled from \(\{0.2, 0.4, 0.6, 0.8, 1.0\}\) and \(\epsilon\) is sampled from a Gaussian scaled to \(10^{-3}\). This results in a linear buoyancy gradient in the vertical direction with a small perturbation. All other fields are initialized to zero. 

Current PDE surrogates usually struggle since the system is highly chaotic and features a transition between laminar and turbulent regimes. Data is obtained from the Well \citep{ohana2025welllargescalecollectiondiverse, burns2020dedalus}, which includes 2D simulations on a \(512\times128\) grid with buoyancy, pressure, and velocity. We use 100 timesteps with an interval of \(\Delta t=0.5\). Models are evaluated after burn-in, starting from timestep 30 (\(t=15\)). 

\paragraph{PlaSim} Global 3D atmospheric data are generated from an intermediate-complexity climate model (PlaSim) to evaluate emulators for weather forecasting and climate prediction \citep{plasim, pnas_rangone, plasim_2}. PlaSim solves the Navier-Stokes equation on a rotating sphere along with parameterizations for various atmospheric  (e.g., moist convection, radiation) and land processes, while the sea surface temperature and sea ice cover are prescribed and vary with a yearly period. A full description of the climate variables and the pressure levels used in the climate simulation dataset is given in Table \ref{tab:variables}. 

Prognostic atmospheric variables (temperature, humidity, zonal and meridional wind) are saved at a resolution of $(128 \times 64  \times 10)$ (latitude, longitude, model level) on an Gaussian horizontal grid. These are then vertically interpolated onto 13 equipressure levels, and the geopotential height is computed using the hydrostatic equation. 8 surface variables, including 2-meter temperature and accumulated precipitation, and 6 forcing variables are also saved. Models are trained on 100 years of data at 6-hour intervals and validated on a year of held-out data, except when evaluating climatological biases, which uses 10 years of data. This results in \(\sim\)144,000 training samples and \(\sim\)1,440 validation samples. 

\begin{table}[t!]
\footnotesize
\centering
\caption{Climate variables grouped into surface, atmospheric, and forcing variables. Additionally, 13 pressure levels are used.}
\newcommand{\var}[2]{#2 (\texttt{#1})}
\addtolength{\tabcolsep}{-0.4em}
\begin{tabular}{llll}
\toprule
\textbf{Surface Variables (8)} & \textbf{Atmospheric Variables (5)} & \textbf{Forcing Variables (6)} & \textbf{Pressure Levels (13)} \\
\midrule
\begin{tabular}[t]{@{}l@{}}
\var{evap}{lwe of water evaporation} \\
\var{mrro}{surface runoff} \\
\var{mrso}{lwe of soil moisture content} \\
\var{pl}{log surface pressure} \\
\var{pr\_12h}{12h acc. precipitation} \\
\var{pr\_6h}{6h acc. precipitation} \\
\var{t2m}{air temperature 2m} \\
\var{ts}{surface temperature}
\end{tabular}
&
\begin{tabular}[t]{@{}l@{}}
\var{hus}{specific humidity} \\
\var{ta}{air temperature} \\
\var{ua}{eastward wind} \\
\var{va}{northward wind} \\
\var{zg}{geopotential}
\end{tabular}
&
\begin{tabular}[t]{@{}l@{}}
\var{lsm}{land sea mask} \\
\var{sg}{surface geopotential} \\
\var{z0}{surface roughness length} \\
\var{rsdt}{TOA Incident Radiation} \\
\var{sic}{sea ice cover} \\
\var{sst}{sea surface temperature}
\end{tabular}
&
\begin{tabular}[t]{@{}l@{}}
50, 100, 150, 200,\\
250, 300, 400, 500,\\
600, 700, 850, 925,\\
1000
\end{tabular}
\\
\bottomrule
\end{tabular}
\label{tab:variables}
\end{table}

\subsection{Metrics} 
\paragraph{Deterministic Metrics} We follow standard metrics to evaluate the deterministic and statistical performance of surrogate models. For PDE tasks, we use Variance Scaled RMSE (VRMSE), and for weather forecasting, we use latitude-weighted RMSE (lRMSE). Given a spatio-temporal data sample \(\mathbf{u} \in \mathbb{R}^{n_t\times n_x \times n_y}\) and a model rollout \(\mathbf{u}_\theta \in \mathbb{R}^{n_t\times n_x \times n_y}\) the Variance-Scaled RMSE (VRMSE) is given by \citep{ohana2025welllargescalecollectiondiverse}:
\begin{equation}
    VRMSE(\mathbf{u}, \mathbf{u}_\theta) = \frac{1}{n_t} \sum _{t=1}^{n_t}\frac{||\mathbf{u}_\theta(t) - \mathbf{u}(t)||_2}{||\mathbf{u}(t) - \bar{\mathbf{u}}(t)||_2 + \epsilon}, \quad ||\mathbf{u}||_2 = \sqrt{\frac{1}{n_xn_y}\sum_{i=1}^{n_x}\sum_{j=1}^{n_y}|\mathbf{u}(i, j)|^2}
\end{equation}
The term \(\epsilon = 10^{-6}\) is added for numerical stability. This metric scales the error by the variance of the input sample, then averages over the prediction horizon \(n_t\). Predicting the mean field will result in \(VRMSE(u, \bar u) \approx 1\), which is a useful interpretation. 

Latitude-weighted RMSE (lRMSE) is calculated at each lead time, for each variable and pressure level separately. For a forecasted variable \(f_{tl} \in \mathbb{R}^{n_{lat}\times n_{lon}}\) and ground-truth \(o_{tl} \in \mathbb{R}^{n_{lat}\times n_{lon}}\) at level \(l\) and time \(t\), lRMSE is given by \citet{rasp2024weatherbench2benchmarkgeneration}:
\begin{equation}
    lRMSE = \sqrt{\frac{1}{n_{lat} n_{lon}}\sum _{i=1}^{n_{lat}}\sum_{j=1}^{n_{lon}} w(i)(f_{tl}(i, j) - o_{tl}(i, j))^2}, \quad w(i) = \frac{\sin \theta ^u_i - \sin \theta_i^l}{\frac{1}{I}\sum _{i=1}^I(\sin \theta _i ^u - \sin \theta _i ^l)}
\end{equation}
For latitude weights \(w(i)\), \(i\) denotes the index of the discretized latitude, and \(\theta_i^u\) and \(\theta_i^l\) denote the upper and lower bounds of the cell at latitude \(i\). Latitude weights are used to account for distortion at the poles in an equiangular grid, which would otherwise over-emphasize predicted values near the poles. 


Annual climatological biases are also calculated to evaluate the consistency of the climate emulator; these are calculated as the lRMSE between the time average of a 10-year ground truth from PlaSim and a 10-year emulation for each atmospheric and surface variable. In particular, the 10-year bias is calculated for each variable and pressure level by: 
\begin{equation}
    \text{Bias}_{\text{10-year}} = lRMSE(f_{avg},  o_{avg}), \quad f_{avg} = \frac{1}{n_t}\sum_{t=1}^{n_t}f_{l}(t)
\end{equation}
where \(n_t \approx 10*365*4\), since a forecast is made every 6 hours for each day in 10 years. Note that \(f_l(t) \in \mathbb{R}^{n_{lat}\times n_{lon}}\), \(f_{avg} \in \mathbb{R}^{n_{lat}\times n_{lon}}\), and lRMSE reduces over latitude and longitude. 

\paragraph{Statistical Metrics} For PDE tasks, the statistical consistency of fluid flows is evaluated by comparing the power spectrum of predicted and true rollouts with spectral RMSE (SRMSE). Exact predictions of turbulent flows over time is usually not possible for PDE surrogates, therefore spectral metrics can quantify the distribution and scale of predicted features rather than relying on point-wise accuracy. 

Given a spatio-temporal data sample \(\mathbf{u} \in \mathbb{R}^{n_t\times n_x \times n_y}\) and a model rollout \(\mathbf{u}_\theta \in \mathbb{R}^{n_t\times n_x \times n_y}\) the DFT is used at each timestep to generate the power spectrum \(p(t)\) and frequencies \(k\). The power spectrum is partitioned based on its frequency into three evenly log-spaced bins (low, mid, high). Following \citet{ohana2025welllargescalecollectiondiverse}, the SRMSE between the true and predicted spectra (\(p, p_\theta\)) for each bin is then calculated as:
\begin{equation}
    SRMSE(p, p_\theta) =  \frac{1}{n_t} \sum _{t=1}^{n_t}\sqrt{1 - \left(\frac{p(t)}{p_\theta(t)}\right)^2}
\end{equation}
For inputs with multiple channels, the SRMSE is calculated for each channel separately and then averaged. 

\paragraph{Probabilistic Metrics} For probabilistic forecasting, the performance of an ensemble of predictions is measured using the continuous ranked probability score (CRPS) and the spread-skill ratio (SSR). CRPS is minimized when samples from the generative model are drawn from the same distribution as the data \citep{zamo_crps}. Similar to lRMSE, CRPS is calculated at each lead time, and for each variable and pressure level separately, however with the addition of \(M\) different ensemble members. CRPS makes use of the latitude-weighted mean absolute error (lMAE):
\begin{equation}
    lMAE = \frac{1}{n_{lat} n_{lon}}\sum _{i=1}^{n_{lat}}\sum_{j=1}^{n_{lon}} w(i)|f_{tl}(i, j) - o_{tl}(i, j)|
\end{equation}
CRPS for a given lead time \(t\), variable, and level \(l\) is given by:
\begin{equation}
    CRPS_{tl} = \frac{1}{M}\sum_{m=1}^{M} lMAE(f_{tl}^{(m)} , o_{tl}) - \frac{1}{2M(M-1)}\sum_{m=1}^{M}\sum_{n=1}^N lMAE(f^{(m)}_{tl}, f^{(n)}_{tl})
\end{equation}
The first term penalizes deviations of the individual ensemble members from the ground truth, and the second term encourages spread between ensemble members.

For the spread-skill ratio (SSR), the spread is calculated for a given time, level, and variable as the square root of the ensemble variance:
\begin{equation}
    \text{Spread}_{tl} = \sqrt{\sum _{i=1}^{n_{lat}}\sum_{j=1}^{n_{lon}} w(i)var_m(f_{tl}^{(m)}(i, j))}
\end{equation}
where \(var_m\) calculates the variance over \(M\) ensemble members. The ensemble skill is given by the RMSE of the ensemble mean:
\begin{equation}
    \bar{f}_{tl} = \frac{1}{M}\sum_{m=1}^{M}f^{(m)}_{tl}, \quad \text{Skill}_{tl} = lRMSE(\bar{f}_{tl}, o_{tl})
\end{equation}
where \(f^{(m)}_{tl}\in\mathbb{R}^{n_{lat}\times n_{lon}}\) is an individual forecast for a variable at lead time \(t\) and level \(l\). The SSR is given by: 
\begin{equation}
    SSR_{tl} = \frac{\text{Spread}_{tl}}{\text{Skill}_{tl}}
\end{equation}
SSR values of 1 are considered optimal \citep{zamo_crps} as the uncertainty of the emulator matches its error. Calculating SSR for a given lead time is done similarly to CRPS, where SSR values for a lead time are averaged across ensemble forecasts made for the validation year. SSR values less than 1 indicate an under-dispersive forecast, where the ensemble fails to capture the full range of possible outcomes, while values over 1 indicate an over-dispersive forecast, where the ensemble is overly uncertain. For all evaluations, an ensemble size of 8 is used.

\section{Results \label{sec:results}}
\begin{table}[t]
    \centering
    \resizebox{\linewidth}{!}{%
    \begin{tabular}{l l | c c c c | c c c}
        \toprule
        Model & NFE & VRMSE & SRMSE$_L$ & SRMSE$_M$ & SRMSE$_H$ & SSR & CRPS & Ens.\ VRMSE \\
        \midrule
        FNO             & 1   & 0.589\footnotesize$\pm$0.009 & \cellcolor{mygray}0.079\footnotesize$\pm$0.009 & \cellcolor{mygray}0.046\footnotesize$\pm$0.003 & \cellcolor{mygray}0.406\footnotesize$\pm$0.013 & --- & 1.507\footnotesize$\pm$0.028 & 0.589\footnotesize$\pm$0.009 \\
        DDPM            & 100 & 0.539\footnotesize$\pm$0.002 & 0.125\footnotesize$\pm$0.020 & 0.076\footnotesize$\pm$0.004 & 0.795\footnotesize$\pm$0.000 & \cellcolor{mygray2}0.937\footnotesize$\pm$0.018 & 0.707\footnotesize$\pm$0.008 & 0.413\footnotesize$\pm$0.003 \\
        DDIM            & 10  & 0.604\footnotesize$\pm$0.004 & 0.275\footnotesize$\pm$0.020 & 0.170\footnotesize$\pm$0.002 & 0.801\footnotesize$\pm$0.001 & 0.355\footnotesize$\pm$0.007 & 1.180\footnotesize$\pm$0.009 & 0.557\footnotesize$\pm$0.003 \\
        TSM             & 10  & 1.914\footnotesize$\pm$1.088 & 0.399\footnotesize$\pm$0.072 & 0.463\footnotesize$\pm$0.151 & 0.891\footnotesize$\pm$0.039 & 0.748\footnotesize$\pm$0.144 & 3.343\footnotesize$\pm$2.206 & 1.576\footnotesize$\pm$0.892 \\
        EDM (Det.)      & 10  & \cellcolor{mygray2}0.426\footnotesize$\pm$0.002 & 0.100\footnotesize$\pm$0.005 & 0.064\footnotesize$\pm$0.002 & \cellcolor{mygray2}0.793\footnotesize$\pm$0.000 & 0.744\footnotesize$\pm$0.025 & \cellcolor{mygray2}0.583\footnotesize$\pm$0.013 & \cellcolor{mygray2}0.340\footnotesize$\pm$0.005 \\
        EDM (Sto.)      & 20  & 0.427\footnotesize$\pm$0.001 & 0.088\footnotesize$\pm$0.013 & 0.061\footnotesize$\pm$0.002 & \cellcolor{mygray2}0.793\footnotesize$\pm$0.001 & 0.720\footnotesize$\pm$0.016 & 0.598\footnotesize$\pm$0.004 & 0.346\footnotesize$\pm$0.004 \\
        FM              & 10  & 0.435\footnotesize$\pm$0.018 & 0.097\footnotesize$\pm$0.011 & 0.065\footnotesize$\pm$0.003 & 0.794\footnotesize$\pm$0.001 & 0.394\footnotesize$\pm$0.022 & 0.754\footnotesize$\pm$0.051 & 0.388\footnotesize$\pm$0.021 \\
        SI (Det.) & 10  & \cellcolor{mygray}0.391\footnotesize$\pm$0.009 & \cellcolor{mygray2}0.086\footnotesize$\pm$0.004 & \cellcolor{mygray2}0.055\footnotesize$\pm$0.001 & \cellcolor{mygray2}0.793\footnotesize$\pm$0.000 & --- & 0.948\footnotesize$\pm$0.022 & 0.391\footnotesize$\pm$0.009 \\
        SI (Sto.) & 20  & 0.441\footnotesize$\pm$0.013 & 0.095\footnotesize$\pm$0.014 & 0.064\footnotesize$\pm$0.003 & \cellcolor{mygray2}0.793\footnotesize$\pm$0.000 & \cellcolor{mygray}0.993\footnotesize$\pm$0.012 & \cellcolor{mygray}0.543\footnotesize$\pm$0.011 & \cellcolor{mygray}0.329\footnotesize$\pm$0.006 \\
        \midrule
        AE              & 1   & 0.011 & 0.011 & 0.006 & 0.791 & --- & --- & --- \\
        \bottomrule
    \end{tabular}}
    \caption{Pointwise (VRMSE), spectral (SRMSE), and probabilistic errors on Kolmogorov Flow. Errors are reported as mean $\pm$ standard deviation over three seeds. $L$, $M$, $H$ denote low-, mid-, and high-band SRMSE.
    VRMSE and SRMSE are computed for one ensemble member; SSR, CRPS, and Ens.\ VRMSE (VRMSE of the ensemble mean) use 8-member ensembles.
    SSR is undefined (---) for deterministic models, whose Ens.\ VRMSE is that of their single forecast.}
    \label{tab:km_flow}
\end{table}

\begin{figure}[t]
    \centering
    \includegraphics[width=\linewidth]{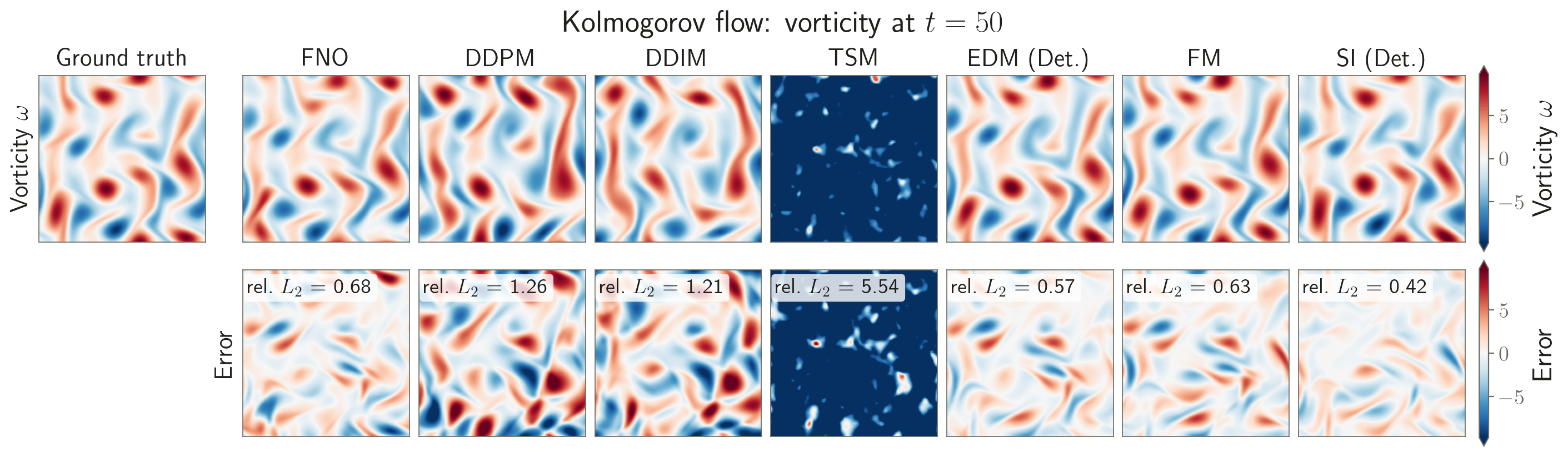}
    \caption{Model predictions of Kolmogorov Flow at timestep \(t=50\), for a single rollout. Deterministic versions of EDM and SI are plotted, and the relative L2 error between predicted and true frames are shown. }
    \label{fig:km_flow_plots}
\end{figure}

In this section, we present results after training each generative model on the three physical systems. Within each task, the model architectures, the training compute and the size of the dataset are kept constant. Each system also exhibits different dynamics and metrics of interest, resulting in different insights for each set of experiments. For clarity, the main results evaluate models at a fixed number of function evaluations (NFEs); additional results that sweep across NFEs are shown in Appendix \ref{app:nfes}. 

\subsection{Kolmogorov Flow}

We report the performance of the models is shown in Table \ref{tab:km_flow}, where the lowest errors are shaded and the second-lowest errors are lightly shaded. All metrics are reported as mean \(\pm\) standard deviation across three random seeds. The number of function evaluations (NFEs) needed for a single prediction is shown, and the autoencoder (AE) performance is given as a reference for the upper bound on performance. The NFEs are standardized to 10 for deterministic samplers and 20 for stochastic samplers, with the exception of DDPM. Additionally, model predictions and errors of at \(t=50\) for a single rollout are plotted in Figure \ref{fig:km_flow_plots}. 

We can observe that stochastic interpolants perform well in this problem. SI models achieve the best scores on probabilistic metrics (SSR and CRPS) while also attaining low deterministic error, both for a single sample (VRMSE) and for the ensemble mean (Ens. VRMSE). SI models also achieve low spectral errors, though they fall short of the pixel-space FNO model, likely due to limitations of the autoencoder. FNO approximates the dataset's spectrum well because the Kolmogorov Flow dataset is laminar and mode-truncated, which aligns closely with FNO's truncated Fourier representation. We can additionally observe that in the highest frequency band, the spectral error of the autoencoder also thresholds the spectral error of latent generative models. However, generative models still achieve lower VRMSE compared to FNO, likely due to more stable rollouts, and inherently offer better calibration and probabilistic scores.

There are also broader trends when comparing the performance of different generative frameworks. First, if deterministic accuracy is the priority, averaging an ensemble of generative samples can reduce VRMSE by an additional 10--20\%. Furthermore, contemporary approaches such as EDM and flow matching (FM) generally outperform DDPM/DDIM and are better default choices, although DDPM remains well calibrated (SSR\(\approx\! 1\)). Deterministic samplers also tend to be under-dispersive: DDIM and FM exhibit low SSR and high CRPS, indicating insufficient ensemble spread. EDM is a notable exception, suggesting that it is still a competitive choice for Gaussian denoising models compared to flow matching, which is seen as a more current approach \citep{armegioiu2025rectifiedflowsfastmultiscale, baldan2025flowmatchingmeetspdes}. Lastly, the choice of source distribution seems to also have an effect on performance. By shifting from flow matching (\(x_0 \sim N(0, I)\)) to a stochastic interpolant (\(x_0 \sim \delta(u_t)\)) we observe a modest gain in performance.

Finally, we report additional results in which all models are trained with varying strides ($\Delta t = 0.2, 0.4, 0.8\,\mathrm{s}$), corresponding to 100, 50, and 25 autoregressive steps on Kolmogorov Flow (Appendix \ref{app:stride}, Table \ref{tab:km_flow_stride}). As $\Delta t$ increases, rollout error decreases slightly across VRMSE, SRMSE, and CRPS, and models also become better calibrated (SSR $\approx 1$). We attribute this to the reduced number of autoregressive calls, which limits error accumulation. However, larger strides come at the cost of coarser temporal resolution, which may be undesirable in some applications. We further observe that the relative advantage of SI models diminishes as $\Delta t$ grows, with models such as EDM achieving better calibration and ensemble VRMSE at larger strides. This is consistent with SI models relying on the proximity of consecutive states: as the prediction interval widens its advantage can be reduced.

\subsection{Rayleigh-Bénard Convection} Rayleigh-Bénard Convection offers a more complex system featuring a transition between laminar and turbulent states. We report the performance of the models is shown in Table \ref{tab:rayleigh_benard}, where the lowest errors are shaded and the second-lowest errors are lightly shaded. All metrics are reported as mean \(\pm\) standard deviation across three random seeds. The NFEs are standardized to 10 for deterministic samplers and 50 for stochastic samplers, with the exception of DDPM.

In this dataset, multiscale features and inherent uncertainty motivate the use of generative models, which show a clear advantage over deterministic surrogates. For FNO, we can see that it has high VRMSE and mid/high-band SRMSE values, which can indicate autoregressive instability and over-smoothing; these are issues generative models are designed to mitigate. 

For generative models, Rayleigh-Bénard convection exposes a trade-off between pointwise and spectral accuracy. After convective mixing, exactly predicting the turbulent state becomes intractable, especially over 100 autoregressive steps, so minimizing pointwise error tends to push models toward overly smoothed predictions. Longer sampling or greater stochasticity counteracts this: taking smaller, random steps introduces perturbations that help recover the true spectrum. However, these same perturbations cause predictions to drift further from the exact state, so pointwise error rises even as spectral characteristics are preserved.

\begin{table}[t]
    \centering
    \resizebox{\linewidth}{!}{%
    \begin{tabular}{l l | c c c c | c c c}
        \toprule
        Model & NFE & VRMSE & SRMSE$_L$ & SRMSE$_M$ & SRMSE$_H$ & SSR & CRPS & Ens.\ VRMSE \\
        \midrule
        FNO             & 1   & 0.589\footnotesize$\pm$0.065 & 0.176\footnotesize$\pm$0.029 & 3.103\footnotesize$\pm$0.412 & 2.783\footnotesize$\pm$0.261 & --- & 0.097\footnotesize$\pm$0.013 & 0.589\footnotesize$\pm$0.065 \\
        DDPM            & 100 & 0.468\footnotesize$\pm$0.009 & 0.147\footnotesize$\pm$0.009 & \cellcolor{mygray2}0.107\footnotesize$\pm$0.003 & \cellcolor{mygray}1.535\footnotesize$\pm$0.027 & \cellcolor{mygray}1.016\footnotesize$\pm$0.018 & 0.036\footnotesize$\pm$0.000 & 0.349\footnotesize$\pm$0.004 \\
        DDIM            & 10  & 0.452\footnotesize$\pm$0.002 & 0.523\footnotesize$\pm$0.010 & 0.713\footnotesize$\pm$0.033 & 2.985\footnotesize$\pm$0.047 & 0.590\footnotesize$\pm$0.008 & 0.049\footnotesize$\pm$0.000 & 0.396\footnotesize$\pm$0.003 \\
        TSM             & 10  & 0.605\footnotesize$\pm$0.045 & 0.348\footnotesize$\pm$0.042 & 0.391\footnotesize$\pm$0.057 & 3.518\footnotesize$\pm$0.165 & 0.577\footnotesize$\pm$0.033 & 0.072\footnotesize$\pm$0.007 & 0.538\footnotesize$\pm$0.037 \\
        EDM (Det.)      & 10  & 0.415\footnotesize$\pm$0.008 & 0.146\footnotesize$\pm$0.003 & 0.167\footnotesize$\pm$0.007 & 2.241\footnotesize$\pm$0.024 & 0.840\footnotesize$\pm$0.002 & \cellcolor{mygray2}0.034\footnotesize$\pm$0.000 & \cellcolor{mygray2}0.329\footnotesize$\pm$0.002 \\
        EDM (Sto.)      & 50  & 0.430\footnotesize$\pm$0.005 & \cellcolor{mygray2}0.130\footnotesize$\pm$0.006 & 0.116\footnotesize$\pm$0.003 & 1.813\footnotesize$\pm$0.014 & 0.921\footnotesize$\pm$0.009 & \cellcolor{mygray2}0.034\footnotesize$\pm$0.000 & \cellcolor{mygray2}0.329\footnotesize$\pm$0.001 \\
        FM              & 10  & 0.419\footnotesize$\pm$0.014 & \cellcolor{mygray2}0.130\footnotesize$\pm$0.002 & 0.148\footnotesize$\pm$0.002 & 2.125\footnotesize$\pm$0.008 & 0.801\footnotesize$\pm$0.016 & 0.035\footnotesize$\pm$0.000 & 0.332\footnotesize$\pm$0.003 \\
        SI (Det.) & 10  & \cellcolor{mygray}0.399\footnotesize$\pm$0.008 & \cellcolor{mygray}0.127\footnotesize$\pm$0.015 & 0.145\footnotesize$\pm$0.004 & 2.566\footnotesize$\pm$0.018 & --- & 0.060\footnotesize$\pm$0.001 & 0.399\footnotesize$\pm$0.008 \\
        SI (Sto.) & 50  & \cellcolor{mygray2}0.414\footnotesize$\pm$0.003 & \cellcolor{mygray}0.127\footnotesize$\pm$0.004 & \cellcolor{mygray}0.096\footnotesize$\pm$0.003 & \cellcolor{mygray2}1.594\footnotesize$\pm$0.008 & \cellcolor{mygray2}0.979\footnotesize$\pm$0.013 & \cellcolor{mygray}0.032\footnotesize$\pm$0.000 & \cellcolor{mygray}0.318\footnotesize$\pm$0.004 \\
        \midrule
        AE              & 1   & 0.027 & 0.086 & 0.061 & 1.528 & --- & --- & --- \\
        \bottomrule
    \end{tabular}}
    \caption{Pointwise (VRMSE), spectral (SRMSE), and probabilistic errors on Rayleigh--B\'enard.
    Errors are reported as mean $\pm$ standard deviation over three seeds. $L$, $M$, $H$ denote low-, mid-, and high-band SRMSE.
    VRMSE and SRMSE are computed per ensemble member; SSR, CRPS, and Ens.\ VRMSE (VRMSE of the ensemble mean) use 8-member ensembles.
    SSR is undefined (---) for deterministic models, whose Ens.\ VRMSE is that of their single forecast.}
    \label{tab:rayleigh_benard}
\end{table}

\begin{figure}[t]
    \centering
    \includegraphics[width=\linewidth]{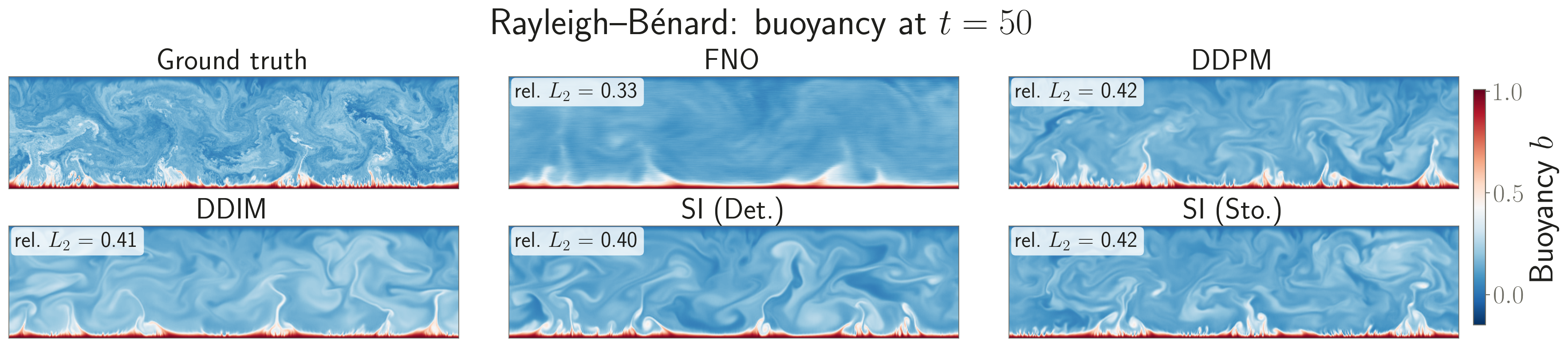}
    \caption{Model predictions of Rayleigh-Bénard Convection at timestep \(t=50\) for a single rollout. For conciseness, only a few models are shown. SI models are plotted with deterministic (Det.) or stochastic (Sto.) sampling. Deterministic surrogates (FNO) tend to over-smooth predictions by minimizing pointwise error, while generative models can resolve finer features when sampled stochastically and  with enough steps.}
    \label{fig:rbc_plots}
\end{figure}

We can see this tradeoff in Table \ref{tab:rayleigh_benard}. Deterministically sampled stochastic interpolants achieve the lowest VRMSE for a single sample but exhibit large mid/high-band spectral errors, while sampling the same model stochastically yields low spectral errors along with good SSR and CRPS scores. DDIM/DDPM and EDM follow the same pattern, favoring either pointwise or spectral accuracy depending on how they are sampled. Figure \ref{fig:rbc_plots} illustrates the trade-off qualitatively, and we can additionally see that a low VRMSE does not generally indicate a good forecast due to over-smoothing. Designing and evaluating generative models for physical emulation therefore requires understanding the metrics of interest and the task at hand, rather than relying on a generic framework. 

Lastly, there are a few trends that are similar to the prior Kolmogorov Flow experiments. Deterministic samplers (DDIM/FM/EDM (Det.)) remain consistently under-dispersive, with lower SSR values than their stochastic counterparts. DDPM also remains well-calibrated and effective at resolving high-frequency spectral content, although it is slow due to needing 100 NFEs. Averaging an ensemble of generative model predictions also reduces VRMSE by 10--20\%, though it reintroduces over-smoothing.

\begin{table}[t]
    \centering
    \resizebox{\linewidth}{!}{%
    \begin{tabular}{l c | c c c c c c c c}
        \toprule
        Var. & Day & SFNO & DDPM & DDIM & EDM (Det.) & EDM (Sto.) & FM & SI (Det.) & SI (Sto.) \\
        & NFE & 1 & 100 & 10 & 10 & 20 & 10 & 10 & 20 \\
        \midrule
        \multirow{4}{*}{$\mathbf{z500}$ [m]} & 1 & 7.8\footnotesize$\pm$0.05 & 6.5\footnotesize$\pm$0.09 & 6.4\footnotesize$\pm$0.06 & \cellcolor{mygray2}6.0\footnotesize$\pm$0.01 & 6.0\footnotesize$\pm$0.01 & 6.1\footnotesize$\pm$0.01 & \cellcolor{mygray}5.9\footnotesize$\pm$0.02 & 6.0\footnotesize$\pm$0.04 \\
         & 3 & 20.6\footnotesize$\pm$0.09 & 13.2\footnotesize$\pm$0.23 & 13.3\footnotesize$\pm$0.17 & 11.8\footnotesize$\pm$0.15 & 11.8\footnotesize$\pm$0.11 & 12.0\footnotesize$\pm$0.14 & \cellcolor{mygray}11.5\footnotesize$\pm$0.12 & \cellcolor{mygray2}11.5\footnotesize$\pm$0.11 \\
         & 5 & 35.6\footnotesize$\pm$0.23 & 22.4\footnotesize$\pm$0.33 & 22.7\footnotesize$\pm$0.32 & 20.4\footnotesize$\pm$0.21 & 20.3\footnotesize$\pm$0.19 & 20.8\footnotesize$\pm$0.32 & \cellcolor{mygray2}20.0\footnotesize$\pm$0.09 & \cellcolor{mygray}19.7\footnotesize$\pm$0.08 \\
         & 10 & 70.7\footnotesize$\pm$1.04 & 46.1\footnotesize$\pm$0.35 & 46.5\footnotesize$\pm$0.18 & 44.1\footnotesize$\pm$0.48 & \cellcolor{mygray2}43.8\footnotesize$\pm$0.25 & 45.2\footnotesize$\pm$0.28 & 49.3\footnotesize$\pm$0.05 & \cellcolor{mygray}42.9\footnotesize$\pm$0.59 \\
        \midrule
        \multirow{4}{*}{$\mathbf{t2m}$ [K]} & 1 & 1.02\footnotesize$\pm$0.007 & 0.74\footnotesize$\pm$0.003 & 0.75\footnotesize$\pm$0.004 & 0.71\footnotesize$\pm$0.000 & 0.71\footnotesize$\pm$0.003 & 0.71\footnotesize$\pm$0.002 & \cellcolor{mygray}0.70\footnotesize$\pm$0.002 & \cellcolor{mygray2}0.70\footnotesize$\pm$0.001 \\
         & 3 & 1.87\footnotesize$\pm$0.022 & 1.05\footnotesize$\pm$0.006 & 1.07\footnotesize$\pm$0.008 & 0.97\footnotesize$\pm$0.002 & 0.97\footnotesize$\pm$0.003 & 0.98\footnotesize$\pm$0.007 & \cellcolor{mygray2}0.97\footnotesize$\pm$0.003 & \cellcolor{mygray}0.96\footnotesize$\pm$0.001 \\
         & 5 & 2.51\footnotesize$\pm$0.042 & 1.41\footnotesize$\pm$0.015 & 1.46\footnotesize$\pm$0.017 & 1.31\footnotesize$\pm$0.011 & \cellcolor{mygray2}1.31\footnotesize$\pm$0.003 & 1.33\footnotesize$\pm$0.015 & 1.34\footnotesize$\pm$0.013 & \cellcolor{mygray}1.28\footnotesize$\pm$0.007 \\
         & 10 & 4.00\footnotesize$\pm$0.130 & 2.31\footnotesize$\pm$0.004 & 2.37\footnotesize$\pm$0.009 & 2.23\footnotesize$\pm$0.023 & \cellcolor{mygray2}2.21\footnotesize$\pm$0.023 & 2.29\footnotesize$\pm$0.021 & 2.56\footnotesize$\pm$0.006 & \cellcolor{mygray}2.19\footnotesize$\pm$0.007 \\
        \midrule
        \multirow{4}{*}{$\mathbf{t850}$ [K]} & 1 & 0.90\footnotesize$\pm$0.005 & 0.81\footnotesize$\pm$0.002 & 0.80\footnotesize$\pm$0.001 & \cellcolor{mygray2}0.78\footnotesize$\pm$0.001 & 0.79\footnotesize$\pm$0.001 & 0.79\footnotesize$\pm$0.003 & \cellcolor{mygray}0.78\footnotesize$\pm$0.000 & 0.79\footnotesize$\pm$0.001 \\
         & 3 & 1.63\footnotesize$\pm$0.022 & 1.09\footnotesize$\pm$0.008 & 1.09\footnotesize$\pm$0.007 & 1.03\footnotesize$\pm$0.004 & 1.03\footnotesize$\pm$0.001 & 1.04\footnotesize$\pm$0.004 & \cellcolor{mygray2}1.02\footnotesize$\pm$0.002 & \cellcolor{mygray}1.02\footnotesize$\pm$0.003 \\
         & 5 & 2.31\footnotesize$\pm$0.033 & 1.49\footnotesize$\pm$0.017 & 1.50\footnotesize$\pm$0.018 & 1.40\footnotesize$\pm$0.008 & \cellcolor{mygray2}1.40\footnotesize$\pm$0.009 & 1.42\footnotesize$\pm$0.018 & 1.41\footnotesize$\pm$0.009 & \cellcolor{mygray}1.37\footnotesize$\pm$0.002 \\
         & 10 & 3.85\footnotesize$\pm$0.045 & 2.48\footnotesize$\pm$0.014 & 2.53\footnotesize$\pm$0.007 & 2.41\footnotesize$\pm$0.020 & \cellcolor{mygray2}2.39\footnotesize$\pm$0.020 & 2.47\footnotesize$\pm$0.010 & 2.76\footnotesize$\pm$0.015 & \cellcolor{mygray}2.36\footnotesize$\pm$0.011 \\
        \midrule
        \multirow{4}{*}{$\mathbf{u250}$ [m/s]} & 1 & \cellcolor{mygray2}1.41\footnotesize$\pm$0.011 & 1.51\footnotesize$\pm$0.007 & 1.50\footnotesize$\pm$0.008 & 1.44\footnotesize$\pm$0.002 & 1.44\footnotesize$\pm$0.003 & 1.45\footnotesize$\pm$0.002 & \cellcolor{mygray}1.41\footnotesize$\pm$0.002 & 1.43\footnotesize$\pm$0.002 \\
         & 3 & 3.26\footnotesize$\pm$0.022 & 2.52\footnotesize$\pm$0.017 & 2.52\footnotesize$\pm$0.024 & 2.35\footnotesize$\pm$0.011 & 2.35\footnotesize$\pm$0.012 & 2.38\footnotesize$\pm$0.013 & \cellcolor{mygray}2.30\footnotesize$\pm$0.008 & \cellcolor{mygray2}2.31\footnotesize$\pm$0.009 \\
         & 5 & 5.06\footnotesize$\pm$0.004 & 3.59\footnotesize$\pm$0.029 & 3.60\footnotesize$\pm$0.033 & 3.36\footnotesize$\pm$0.039 & 3.36\footnotesize$\pm$0.034 & 3.41\footnotesize$\pm$0.024 & \cellcolor{mygray2}3.32\footnotesize$\pm$0.015 & \cellcolor{mygray}3.25\footnotesize$\pm$0.006 \\
         & 10 & 9.20\footnotesize$\pm$0.130 & 6.16\footnotesize$\pm$0.048 & 6.24\footnotesize$\pm$0.033 & 5.97\footnotesize$\pm$0.035 & \cellcolor{mygray2}5.92\footnotesize$\pm$0.019 & 6.05\footnotesize$\pm$0.046 & 6.62\footnotesize$\pm$0.008 & \cellcolor{mygray}5.78\footnotesize$\pm$0.062 \\
        \midrule
        \multirow{4}{*}{$\mathbf{pr\_6h}$ [mm]} & 1 & 0.68\footnotesize$\pm$0.004 & 0.63\footnotesize$\pm$0.001 & 0.63\footnotesize$\pm$0.002 & 0.59\footnotesize$\pm$0.001 & 0.59\footnotesize$\pm$0.001 & \cellcolor{mygray2}0.59\footnotesize$\pm$0.001 & \cellcolor{mygray}0.58\footnotesize$\pm$0.001 & 0.59\footnotesize$\pm$0.001 \\
         & 3 & 1.05\footnotesize$\pm$0.007 & 0.85\footnotesize$\pm$0.006 & 0.86\footnotesize$\pm$0.005 & \cellcolor{mygray2}0.82\footnotesize$\pm$0.002 & 0.82\footnotesize$\pm$0.003 & 0.82\footnotesize$\pm$0.003 & 0.83\footnotesize$\pm$0.002 & \cellcolor{mygray}0.80\footnotesize$\pm$0.004 \\
         & 5 & 1.30\footnotesize$\pm$0.005 & 1.00\footnotesize$\pm$0.004 & 1.01\footnotesize$\pm$0.005 & 0.99\footnotesize$\pm$0.004 & \cellcolor{mygray2}0.98\footnotesize$\pm$0.006 & 0.99\footnotesize$\pm$0.006 & 1.05\footnotesize$\pm$0.007 & \cellcolor{mygray}0.96\footnotesize$\pm$0.001 \\
         & 10 & 1.69\footnotesize$\pm$0.074 & 1.18\footnotesize$\pm$0.001 & 1.18\footnotesize$\pm$0.005 & 1.18\footnotesize$\pm$0.004 & \cellcolor{mygray2}1.17\footnotesize$\pm$0.003 & 1.18\footnotesize$\pm$0.006 & 1.40\footnotesize$\pm$0.003 & \cellcolor{mygray}1.16\footnotesize$\pm$0.003 \\
        \bottomrule
    \end{tabular}}
    \caption{Latitude-weighted RMSE of the ensemble mean on weather forecasting, at lead times of 1, 3, 5 and 10 days. Mean $\pm$ standard deviation over three seeds. Stochastic models use 8-member ensembles; SFNO and SI ($\sigma=0$) are deterministic, so theirs is the RMSE of their single forecast.}
    \label{tab:climate_ens_lrmse}
\end{table}

\begin{figure*}
    \centering
    \includegraphics[width=\linewidth]{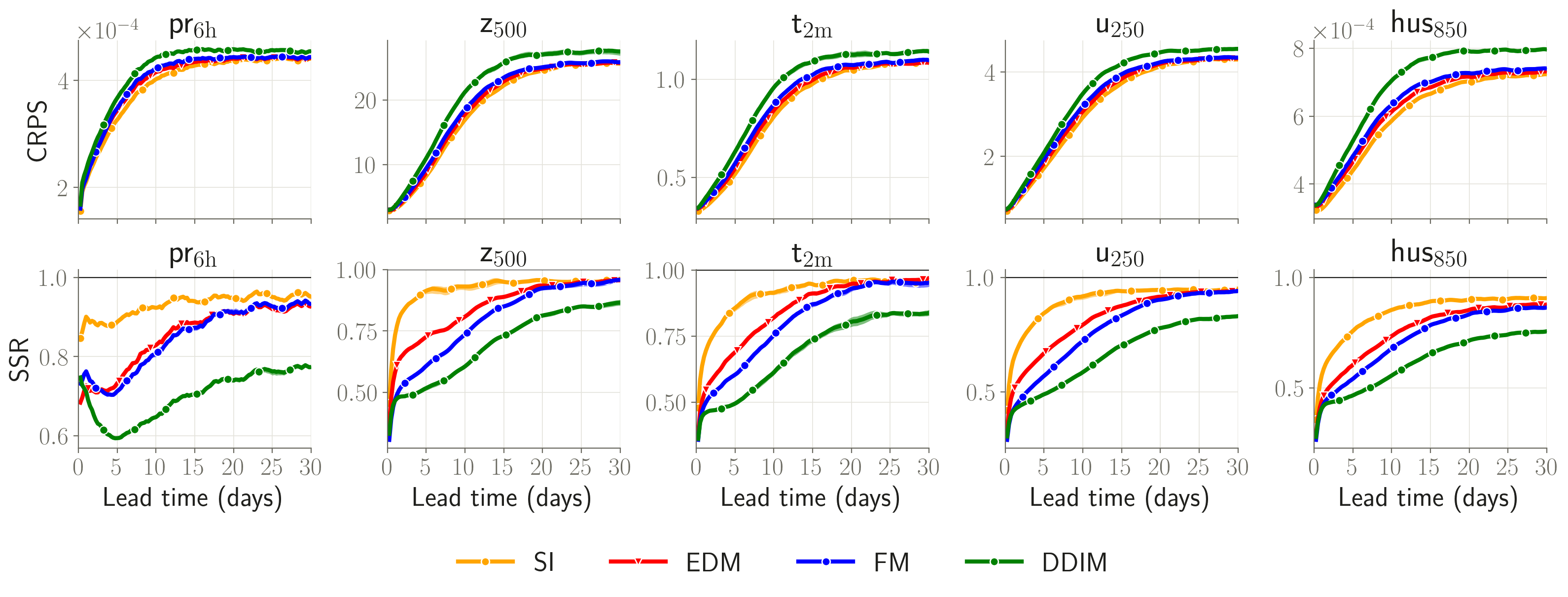}
    \caption{CRPS and SSR plots for generative models over 30 days. DDIM and FM use 10 ODE sampling steps, while EDM and SI use 20 SDE sampling steps. Lines show the mean over three random seeds; shaded regions span the min–max range across seeds, which is small. }
    \label{fig:crps_ssr}
\end{figure*}

\subsection{Weather Forecasting} Many commonly studied PDEs are deterministic; when fully observed, a future state should be known provided that a sufficiently small time interval is used. Although uncertainty and statistical metrics are useful for PDEs, climate systems benefit more directly from probabilistic modeling due to the need for well-calibrated forecasts as well as inherent uncertainty in weather data collection and numerical weather prediction \citep{palmer2019stochastic,ChallengesandBenchmarkDatasetsforMachineLearningintheAtmosphericSciencesDefinitionStatusandOutlook,bracco2025machine}. This makes weather forecasting a good benchmark not only for evaluating the accuracy of generative models but also for understanding their ability to approximate underlying distributions and to capture uncertainty. 

After training, models are evaluated on medium-range weather forecasting for up to 10 days. Following standard practice in weather forecasting, we report the latitude-weighted RMSE (lRMSE) of the ensemble mean, in Table \ref{tab:climate_ens_lrmse}. SFNO and SI (Det.) models report the lRMSE of a single rollout. The NFEs are set to 10 for deterministic samplers and 20 for stochastic samplers, with the exception of DDPM. Overall, we can see that SI models tend to perform well in deterministic accuracy. At short lead times, ensembling offers little benefit and deterministic samplers perform well. At longer lead times, as the future becomes more uncertain, stochastic samplers become advantageous, and their ensemble VRMSE improves relative to deterministic samplers. Again, we can see that EDM and flow matching are better default choices for Gaussian denoising models compared to DDPM and in this case, EDM can even outperform flow matching. 

Although low lRMSE is associated with a good forecast, it has many shortcomings such as encouraging overly smoothed predictions, and minimizing lRMSE can produce forecasts that lack physical consistency or fidelity \cite{bonavita}. Therefore, to better understand the performance of generative models, we also examine the CRPS and SSR of ensemble forecasts. Ensembles are initialized at every 5th day in the validation year to make a 30-day forecast; the CRPS/SSR is calculated at each timestep and metrics are averaged across all initializations. The resulting CRPS and SSR plots are shown in Figure \ref{fig:crps_ssr}. DDPM is omitted due to its computational expense and its lower performance on 10-day forecasts. 

Up to 30 days, stochastic interpolants show lower CRPS and better-calibrated SSR values. Furthermore, EDM has slightly better calibration and CRPS compared to flow matching. While most generative models are under-dispersive at shorter lead times, SI models tend to mitigate this and calibrate their uncertainty earlier. Lastly, the spread between different seeds is quite small, which suggests that the model initialization does not significantly affect the probabilistic performance of weather models. 

\begin{table}[t]
    \centering
    \begin{tabular}{lcccccc}
        \toprule
        Var: & $\mathbf{z500}$ [m]& $\mathbf{t2m}$ [K] & $\mathbf{t850}$ [K]& $\mathbf{u250}$ [m/s] & $\mathbf{pr\_6h}$ [mm] & $\mathbf{hus850}$ [g/kg] \\
        \midrule
        DDIM & 9.77{\scriptsize\,$\pm$\,1.04} & 0.53{\scriptsize\,$\pm$\,0.14} & 0.45{\scriptsize\,$\pm$\,0.13} & 1.33{\scriptsize\,$\pm$\,0.15} & 0.106{\scriptsize\,$\pm$\,0.001} & 0.426{\scriptsize\,$\pm$\,0.015} \\
        EDM\footnotesize(Sto.) & 4.31{\scriptsize\,$\pm$\,0.71} & 0.34{\scriptsize\,$\pm$\,0.01} & 0.28{\scriptsize\,$\pm$\,0.01} & 0.63{\scriptsize\,$\pm$\,0.11} & 0.069{\scriptsize\,$\pm$\,0.006} & 0.270{\scriptsize\,$\pm$\,0.009} \\
        FM & \cellcolor{mygray2}4.18{\scriptsize\,$\pm$\,0.71} & \cellcolor{mygray2}0.28{\scriptsize\,$\pm$\,0.01} & \cellcolor{mygray2}0.25{\scriptsize\,$\pm$\,0.02} & \cellcolor{mygray2}0.55{\scriptsize\,$\pm$\,0.06} & \cellcolor{mygray2}0.062{\scriptsize\,$\pm$\,0.002} & \cellcolor{mygray2}0.202{\scriptsize\,$\pm$\,0.017} \\
        SI\footnotesize(Sto.) & \cellcolor{mygray}3.95{\scriptsize\,$\pm$\,0.30} & \cellcolor{mygray}0.25{\scriptsize\,$\pm$\,0.01} & \cellcolor{mygray}0.23{\scriptsize\,$\pm$\,0.00} & \cellcolor{mygray}0.49{\scriptsize\,$\pm$\,0.02} & \cellcolor{mygray}0.048{\scriptsize\,$\pm$\,0.001} & \cellcolor{mygray}0.128{\scriptsize\,$\pm$\,0.006} \\
        \bottomrule
        
    \end{tabular}
    \caption{10-year climatological biases (latitude-weighted RMSE between the 10-year mean of an autoregressive rollout and the true 10-year mean over the same period). All models use 10 sampling steps. The mean $\pm$ standard deviation over 3 seeds is reported.}
    \label{tab:bias}
\end{table}

\begin{figure*}[t]
    \centering
    \includegraphics[width=\linewidth]{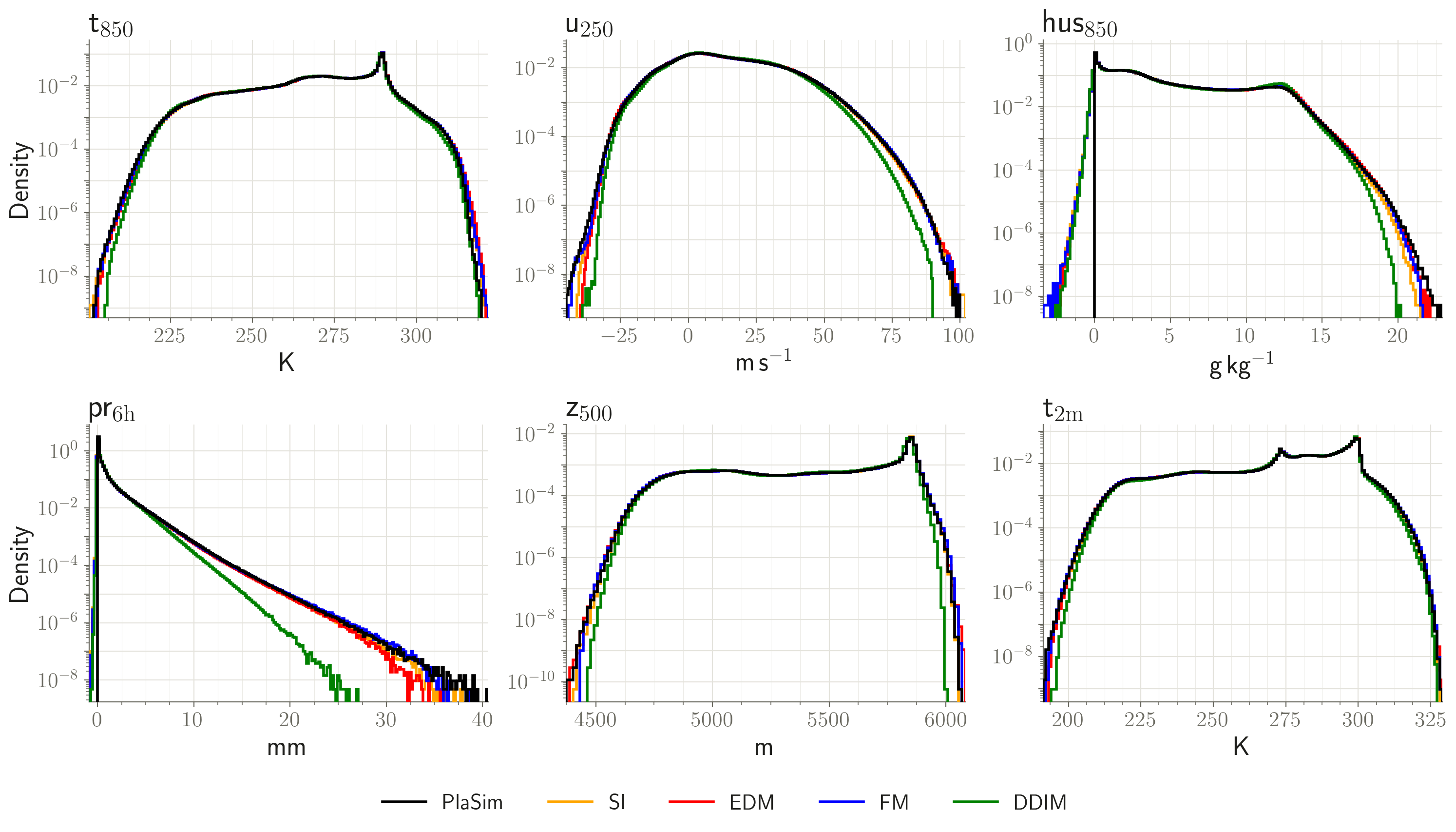}
    \caption{Histograms showing the distributions of climate variables over a 100-year autoregressive rollout (146,095 6-hourly steps), compared to the PlaSim ground truth over the same period. DDIM, EDM, FM, and SI models use 10 sampling steps (identical to Table \ref{tab:bias}). Densities are on a log scale.}
    \label{fig:100yr}
\end{figure*}

\subsection{Climate Emulation} 

An advantage of using PlaSim instead of real-world data is the availability of a very large dataset of true samples from a long PlaSim simulation. This allows models to be evaluated for long-term climate consistency. After training to predict states at 6-hour intervals, generative models are queried to make predictions up to 10 years. For each variable and at each grid point and pressure level, predictions are averaged across all timesteps; the lRMSE between true and predicted averages is the 10-year climatological bias. Prior work has observed the lack of correlation between medium-range forecast error and climatological biases, stemming from error accumulation as the model trained for fast, weather dynamics is used for climatological timescales  \citep{chattopadhyay2023long,cachay2024probabilisticemulationglobalclimate, Watt-Meyer2025}; therefore this metric is a more difficult measure of long-term consistency. Despite this, generative models can be well-suited for long, autoregressive rollouts due to their inherent stability and probabilistic nature. 

The 10-year model biases are reported in Table \ref{tab:bias}, where each model uses 10 sampling steps and the mean \(\pm\) standard deviation over three seeds is reported. SI models tend to do well compared to other frameworks, suggesting that this change in the source distribution has helped to learn a more consistent emulator. However, the spread across seeds is larger than for the medium-range metrics, and the climatological bias fluctuates more from epoch to epoch, although using an exponential moving average (EMA) of the weights mitigates this to some extent. Overall, the low bias is a promising result for SI models: despite being trained on 6-hour steps, they produce autoregressive rollouts of up to 10 years without drifting significantly from the ground-truth climatology.

Lastly, we evaluate long-term emulator behavior by comparing the distributions of climate variables over a 100-year rollout. Following \citet{singh2026evaluatingskillstabilityarchesweather}, each generative model is run for 146,095 six-hourly steps, and pixel-wise values are pooled to estimate densities, which we compare against those of the PlaSim reference trajectory (Figure \ref{fig:100yr}). First, we observe that all generative models remain stable over the full century, which is encouraging for their use as autoregressive emulators. Second, with the exception of DDIM, the models tend to reproduce the reference distributions, including their tails. DDIM is noticeably under-dispersive and fails to capture the tails of variables such as precipitation and humidity, while differences among the remaining models are small. This is promising for generative emulators because the tails represent rare events that may occur only once a year or once a decade, yet generative models can capture them despite being trained only on 6-hour transitions. Lastly, the generative models occasionally make negative predictions of precipitation and humidity, which is non-physical.

\section{Discussion \label{sec:discussion}}

What is the best generative model for a given PDE task? Throughout this work, empirical evidence suggests that this can be dependent on the phenomena and physics of interest. Even when fixing the physical system, the most suitable generative model can also depend on the metric or task of interest. The relative benefit of different generative frameworks can change when evaluating different errors (pointwise vs. spectral), timescales (medium-range vs. climatological), or even variables (temperature vs. precipitation). Despite this, stochastic interpolants have shown promise as a generative model for modeling physical systems. The key inductive bias that is leveraged is the assumption that successive states are related over a given time interval. When architectures and parameterizations are kept constant, empirical evidence suggests that this can be an effective way to train and sample a generative model. Depending on the sampling strategy, stochastic interpolants can achieve low pointwise, spectral, and probabilistic errors, and they perform consistently well across a range of physical systems and tasks.

\begin{figure}[t]
    \centering
    \includegraphics[width=\linewidth]{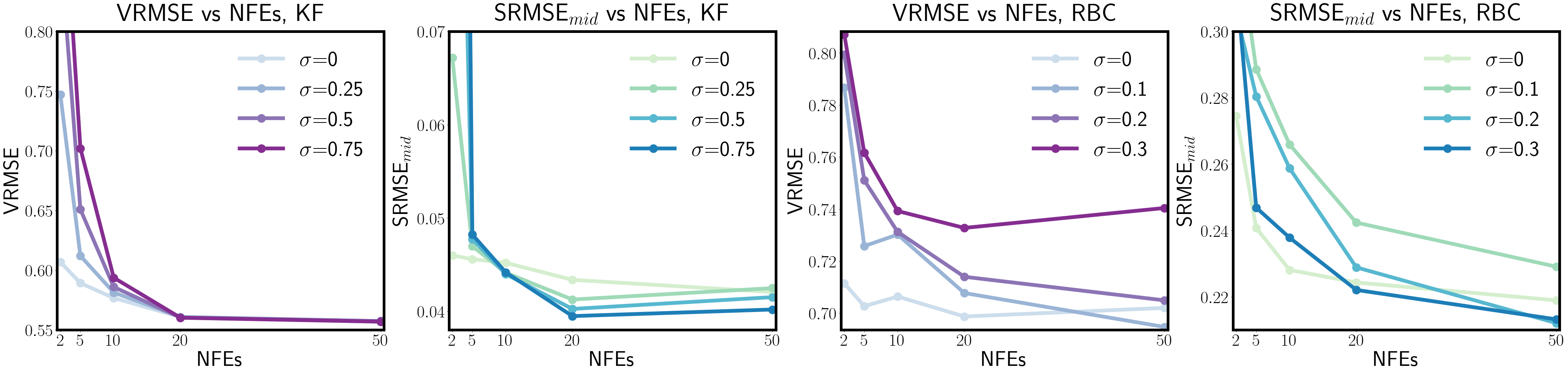}
    \caption{SI-EM models trained on either Kolmogorov Flow (KF) or Rayleigh-Bénard Convection (RBC), sampled with varying stochasticity. VRMSE and SRMSE$_{mid}$ are reported for different numbers of sampling steps (NFEs), and \(\sigma\) scales the noise term \(\sigma(1-t)\).  }
    \label{fig:nfe}
\end{figure}

To better understand the sampling of SI models, we vary the diffusion magnitude during sampling as well as the number of sampling steps in Kolmogorov Flow or Rayleigh-Bénard Convection, and plot the resulting VRMSE and mid-band SRMSE in Figure \ref{fig:nfe}. The trends are consistent with prior intuition. SDEs are more sensitive to discretization error, and at low NFEs, stochasticity harms performance because samplers need more steps to converge. However, when given enough sampling steps, using more stochasticity during sampling usually converges in terms of deterministic accuracy and outperforms less stochastic samplers in terms of spectral accuracy. In more chaotic systems, where the latent space has more variance, excessive noise during sampling can cause predictions to diverge more quickly. However, stochasticity can still be important in recovering the correct spectrum, provided enough sampling steps are used to mix and sample correctly from the underlying modes. 

Beyond interpolants, there are interesting trends for generative models as a whole. EDM and flow matching consistently outperform DDPM while requiring fewer sampling steps. DDPM nonetheless remains a strong choice for resolving high-frequency content and producing well-calibrated uncertainty estimates through its spread-skill ratio. For Gaussian denoising models, we also observe a tradeoff between pointwise and spectral errors that depends on whether sampling is deterministic (ODE) or stochastic (SDE). Finally, we confirm that generative models remain stable during autoregressive rollout, resolve high-frequency content better than deterministic models, and can further reduce pointwise error through ensembling.

Finally, to better understand the benchmarked generative models, we provide a set of ablation studies in Appendix \ref{app:ablation}. Consistent with prior evidence \citep{rozet2025lostlatentspaceempirical} we show that the compression ratio of the autoencoder does not influence prediction error substantially. This is because the primary driver of rollout error is autoregressive error accumulation, rather than reconstruction error; however, excessive compression ratios will degrade the spectral fidelity of the predicted solution. We also verify that generative models are more effective in latent space than in pixel space. Previous works have also observed this, with the hypothesis that a compressed latent space has less variance for autoregressive error accumulation and a well-trained decoder may smooth out errors. Lastly, for stochastic interpolants we investigate the effect of varying the noise coefficient during training, rather than sampling. Adding noise to the interpolant encourages modes from the source and target distributions to mix and produces a smoother trajectory through probability space. This has the effect of regularizing intermediate interpolant states \(x_t\) to improve rollout errors during sampling. However, too much noise can transport intermediate distributions too far from the source or target distributions and increase the difficulty of learning a drift or sampling the transport map. We additionally discuss the transport cost associated with predicting spatiotemporal systems, in Appendix \ref{app:transport}.

\section{Conclusion \label{sec:conclusion}} 
\paragraph{Limitations} There is ample room to continue investigating stochastic interpolants for physical emulation, as it is a highly general framework. Firstly, different source distributions beyond the Gaussian and point masses considered in this work may be beneficial. In fact, these distributions can be seen as the limits of a continuum of source distributions defined as partially noised current states. For certain physics problems, there are also other prior distributions that may be relevant, such as the mean flow in Kolmogorov Flow or the climatology in climate emulation. Additionally, we did not thoroughly investigate many training parameters such as the mixing coefficients or loss function, which may benefit from preconditioning or input/output scaling \citep{karras2022elucidatingdesignspacediffusionbased}. It is also likely that sampling the interpolant timestep \(t\) uniformly is not ideal, as intermediate steps along the transport map are more difficult to learn. Lastly, higher-order samplers could also further improve the performance of stochastic interpolants. 

Additionally, there are several contemporary generative modeling frameworks we did not consider. Consistency models are one candidate for future work, as they can further increase sampling speed without significant performance losses \citep{song2023consistencymodels}. These also have different variants \citep{geng2024consistencymodelseasy, song2023improvedtechniquestrainingconsistency} and can also be seen in the broader framework of Flow Maps \citep{boffi2025buildconsistencymodellearning}. There is also some prior work in applying consistency models to weather forecasting \citep{stock2025swiftautoregressiveconsistencymodel}, which are faster to sample while maintaining accuracy. Additionally, Mean Flows \citep{geng2025meanflowsonestepgenerative} and related frameworks \citep{geng2026improvedmeanflowschallenges} are also a promising family of generative models for physical systems. Lastly, there could be future work to evaluate SI models on more challenging systems, such as using ERA5 \citep{rasp2024weatherbench2benchmarkgeneration} instead of PlaSim, or experimental fluid observations instead of simulated trajectories \citep{hu2026realpdebenchbenchmarkcomplexphysical}. We leave these directions to future work. 

\paragraph{Takeaways} Modeling physical phenomena is challenging, as each system exhibits distinct dynamics, variables, and spatial or temporal scales. This variability suggests that the design of generative surrogates should consider the physical system and desired performance metrics. Our experiments support this idea: even with the same architecture and training budget, changes in the transport map and sampling can result in different deterministic and statistical performance. Despite this, stochastic interpolants can be a strong baseline for generative models, motivated by the proximity of subsequent states in autoregressive prediction. Furthermore, its flexibility can be a useful tool to tailor model performance to specific systems, or to investigate new samplers or transport maps for physical emulation, all under a unified framework. 

\bibliography{bib/climate_diffusion, bib/diffusion, bib/pde_diffusion, bib/pde_flow_matching, bib/pdes, bib/stochastic_interp}
\bibliographystyle{tmlr}

\appendix
\newpage 
\section{Ablation Studies}
\label{app:ablation}

\paragraph{Pixel/Latent Space Ablations} 

We train flow matching and stochastic interpolant models in either pixel or latent space to compare their performance. The Normalized RMSE (NRMSE), or Relative L2 error, is reported for the Kolmogorov Flow validation set in Table \ref{tab:pixel_comp}. Model sizes are kept roughly constant, and all models are trained for the same number of epochs. To account for the larger spatial input to the DiT \((160\times 160)\), a patch size of \(8\times 8 \) was used to match the compression ratio of the autoencoder. 

In general, we find that pixel space models have higher errors than latent space models, in addition to being more expensive to train and query. One hypothesis is that latent space models could be more stable, as the latent space has less variance and a well-trained decoder may smooth out errors. Indeed, we find that pixel space models accumulate error faster than latent space models. 

\begin{table}[hbp]
    \centering
    \caption{Comparison of pixel and latent space generative models on Kolmogorov Flow.}
    \begin{tabular}{l c c c c}
        \toprule 
         Metric & FM$_{pixel}$ & FM$_{latent}$ & SI$_{pixel}$ & SI$_{latent}$\\
         \midrule 
           NRMSE & 0.812 & 0.649 & 0.966 & 0.609\\
         \bottomrule 
    \end{tabular}
    \label{tab:pixel_comp}
\end{table}

\paragraph{Compression Ratio Ablations} We compare the effects of training autoencoders with different compression ratios \((64, 256)\) on the reconstruction error of the autoencoder and the rollout error of the generative model. The NRMSE for Rayleigh-Bénard convection is reported in Table \ref{tab:ratio_comp}. A larger compression ratio results in a smaller latent space, increasing the reconstruction error of the autoencoder. Despite having nearly double the reconstruction error, generative models with more aggressive compression only have a modest increase in rollout error, which is consistent with \citet{rozet2025lostlatentspaceempirical}. 

\begin{table}[h!]
    \centering
    \caption{Comparison of NRMSE for models using different compression ratios (\(\div\)) on Rayleigh-Bénard Convection. Reconstruction error is reported for the autoencoder (AE), while rollout error is reported for the generative models.}
    
    \begin{tabular}{l c c c}
        \toprule 
         \(\div\)\ & AE & FM & SI \\
         \midrule 
           64 & 0.0252 & 0.619 & 0.605\\
           256  & 0.0457 & 0.649 & 0.622\\
         \bottomrule 
    \end{tabular}
    \label{tab:ratio_comp}
\end{table}

\paragraph{Noise Coefficients for Interpolants} 
The stochastic interpolant considered in this work is given by: \(x(t) = \alpha(t) x_0 + \beta(t) x_1 + \sigma(t)W_t\). We define \(\sigma(t) = \sigma (1-t)\) as the noise coefficient, where \(\sigma\) controls the scale of the noise in the stochastic process. More noise can encourage mixing between \(x_0\) and \(x_1\), however, too much noise can make the stochastic process more difficult to learn or sample from. We can see this in Table \ref{tab:sigma_comp}, where SI models are trained to learn stochastic processes with different noise scales \(\sigma\) and are sampled with an Euler sampler using 2 steps. In general, \(\sigma\) can be tuned to find an optimal amount of noise for the stochastic process.

\begin{table}[h!]
    \centering
    \caption{Comparison of different \(\sigma\) coefficients for SI models in Kolmogorov Flow. \(\sigma\) scales the amount of noise in the stochastic process.}
    \begin{tabular}{l c c c c}
        \toprule 
         \(\sigma\): & 0.1 & 0.5 & 1 & 3 \\ 
         \midrule 
        NRMSE & 0.702 & 0.632 & 0.609 & 0.803 \\ 
         \bottomrule 
    \end{tabular}
    \label{tab:sigma_comp}
\end{table}

\paragraph{Ablations for DDPM/DDIM/TSM} 

DDPM, DDIM, and TSM all model the same forward process, but sample the reverse process differently. We perform a hyperparameter sweep for these baselines in Table \ref{tab:ddpm_comp}, and use the best configurations in our experiments. We experiment with different noise schedules as well as different start and ending points for the noise schedule by modifying a scaling factor. For the linear schedule, we set \(\beta_{start} = 0.0001\) and \(\beta_{end} = 0.02\) and sweep over scaling factors \(S\) that modify the schedule by multiplying \(S *\beta_{start}\), \(S *\beta_{end}\), following \citet{shehata2025improved}. Additionally, the cosine noise schedule is modified by changing \(s\) in the expression: \(\bar\alpha_t =\frac{f(t)}{f(0)}, f(t)=cos(\frac{t/T+s}{1+s} * \frac{\pi}{2})^2\). Results are shown in Table \ref{tab:ddpm_comp}. 

We find that a cosine schedule degrades performance for DDPM and TSM, however DDIM benefits from the cosine schedule, with the scale having a small effect on performance. Within linear schedules, the scale \(S=400\) seems to work well. Therefore we train DDPM, DDIM, and TSM models with a linear schedule with \(S=400\) to allow a comparison between samplers for diffusion training. 

\begin{table}[h!]
    \centering
    \caption{Comparison of hyperparameters for DDPM/DDIM/TSM models in Kolmogorov Flow.}
    \begin{tabular}{l c c c c c c}
        \toprule 
         Model & Schedule & Scale & VRMSE & SRMSE$_{low}$ & SRMSE$_{mid}$ & SRMSE$_{high}$  \\ 
         \midrule 
         DDPM & Linear & \(S=200\) & 0.691 & 0.137 & 0.074 & \cellcolor{mygray}0.794\\ 
         DDIM & Linear & \(S=200\) & 0.797 & 1.840 & 0.831 & 0.977\\ 
         TSM & Linear & \(S=200\) & 0.969 & 1.413 & 2.082 & 0.932 \\ 
          DDPM & Linear & \(S=400\) & \cellcolor{mygray} 0.655 & 0.085 & \cellcolor{mygray} 0.056 & \cellcolor{mygray}0.794\\ 
         DDIM & Linear & \(S=400\) & 0.744 & 0.284 & 0.208 & 0.804\\ 
         TSM & Linear & \(S=400\) & 0.827 & 1.541 & 0.892 & 0.896 \\ 
          DDPM & Linear & \(S=800\) & 0.691 & \cellcolor{mygray} 0.079 & 0.074 & 0.800\\ 
         DDIM & Linear & \(S=800\) & 0.671 & 0.080 & 0.098 & 0.799\\ 
         TSM & Linear & \(S=800\)  & \(>\)10 & 0.978 & 0.985 & 0.999\\ 
         \midrule 
       DDPM & Cosine & \(s=0.004\) & 7.091 & 0.526 & 0.569 & 0.909 \\ 
         DDIM & Cosine & \(s=0.004\) & 0.677 & 0.276 & 0.187 & 0.801\\ 
         TSM & Cosine & \(s=0.004\) & \(>\)10 & 1.000 & 1.000 & 1.000\\ 
       DDPM & Cosine & \(s=0.008\) & 7.693 & 0.586 & 0.631 & 0.923\\ 
         DDIM & Cosine & \(s=0.008\) & 0.679 & 0.266 & 0.171 & 0.800 \\
         TSM & Cosine & \(s=0.008\) & \(>\)10 & 1.000 & 1.000 & 1.000\\ 
         \bottomrule 
    \end{tabular}
    \label{tab:ddpm_comp}
\end{table}

\paragraph{Ablations for EDM} Elucidated Diffusion Models (EDMs) reparameterize the forward and reverse process to a common form to better optimize their parameters. We experiment with different noise schedules \(\sigma_{min}, \sigma_{max}\) as well as sampling parameters \(\rho\). Furthermore, we examine an SDE-based sampler by using the Euler-Maruyama method with 20 steps. Following \citet{shehata2025improved}, we adopt SDE sampling parameters \(S_{churn}, S_{min}, S_{max}, S_{noise}\) of \(10, 0, 1, 1e6\) based on tuning performed in their work. Results are shown in Table \ref{tab:edm_comp}. 

\begin{table}[h!]
    \centering
        \caption{Comparison of different hyperparameters for EDM models in Kolmogorov Flow.}
    \begin{tabular}{l c c c c c c}
        \toprule 
         Sampler & \(\sigma_{min}, \sigma_{max}\) & \(\rho\)& VRMSE & SRMSE$_{low}$ & SRMSE$_{mid}$ & SRMSE$_{high}$ \\ 
         \midrule 
        ODE & \(0.002, 80\) & \(3\) & 0.616 & 0.084 & 0.062 & 0.793\\ 
        ODE & \(0.02, 100\) & \(3\) & 0.682 & 0.236 & 0.061 & 0.792\\ 
         ODE & \(0.001, 60\) & \(3\) & 0.692 & 0.252 & 0.074 & 0.792\\ 
         ODE & \(0.002, 80\) & \(7\) &\cellcolor{mygray} 0.606 & \cellcolor{mygray}0.073 & 0.047 & 0.792\\ 
        ODE & \(0.02, 100\) & \(7\) & 0.685 & 0.194 & 0.059 & \cellcolor{mygray}0.791\\ 
         ODE & \(0.001, 60\) & \(7\) & 0.688 & 0.241 & 0.069 & 0.792\\ 
         \midrule 
        SDE & \(0.002, 80\) & \(3\) & 0.625 & 0.091 & 0.046 & 0.794\\ 
        SDE & \(0.02, 100\) & \(3\) & 0.677 & 0.208 & 0.054 & 0.792 \\ 
         SDE & \(0.001, 60\) & \(3\) & 0.684 & 0.226 & 0.059 & \cellcolor{mygray}0.791\\ 
         SDE & \(0.002, 80\) & \(7\) & 0.612 & 0.092 & \cellcolor{mygray}0.045 & 0.794 \\ 
        SDE & \(0.02, 100\) & \(7\) & 0.680 & 0.204 & 0.053 & \cellcolor{mygray}0.791\\ 
         SDE & \(0.001, 60\) & \(7\) & 0.682 & 0.218 & 0.061 & 0.792\\ 
         \bottomrule 
    \end{tabular}
    \label{tab:edm_comp}
\end{table}

For simpler systems like Kolmogorov Flow, SDE-based sampling does not seem to help despite using twice the number of sampling steps, except when resolving the mid-band \(SRMSE\). Furthermore, the original noise schedule reported in \citet{karras2022elucidatingdesignspacediffusionbased} seems to work the best with \(\sigma_{min}=0.002, \sigma_{max}=80\), likely since we normalize our data to a similar range. Lastly, there are small changes in performance due to the weighting of the sampling steps, controlled by \(\rho\). Therefore, we opt to use EDM with an ODE sampler and with parameters \(\sigma_{min}=0.002, \sigma_{max}=80, \rho = 7\). 

\section{Additional Details}
\subsection{Generative Models}
\label{app:generative}

All generative frameworks share the DiT backbone, optimizer, and training budget described in Appendix~\ref{app:arch}, and differ only in their training objective and sampler. Each model learns the conditional distribution $p(\mathbf{z}_{t+\Delta t} \mid \mathbf{z}_t)$ in latent space. Below, $\mathbf{x}_0 = \mathbf{z}_t$ is the conditioning state, $\mathbf{x}_1 = \mathbf{z}_{t+\Delta t}$ is the target, $\boldsymbol{\epsilon} \sim \mathcal{N}(0, \mathbf{I})$, and $s$ is the diffusion or interpolant time.

\paragraph{DDPM, DDIM, and TSM.} These three share one trained $\boldsymbol{\epsilon}$-prediction network and differ only in how they sample. We use $N = 100$ diffusion steps with a linear $\beta$ schedule, where $\beta_s$ runs from $4 \times 10^{-4}$ to $8 \times 10^{-2}$. The network is trained with an MSE loss on the added noise, over the full range of diffusion steps. DDPM samples with the full 100-step ancestral chain. DDIM samples deterministically with 10 evenly spaced steps. TSM truncates the ancestral chain: it runs from $s = N$ to $s = \lfloor pN \rfloor$ and then takes a single $\hat{\mathbf{x}}_1$ prediction.

\paragraph{EDM.} We follow the EDM preconditioning and training setup of \citet{karras2022elucidatingdesignspacediffusionbased}, with $\sigma_{\text{data}} = 0.5$ and noise levels drawn as $\ln \sigma \sim \mathcal{N}(P_{\text{mean}}, P_{\text{std}}^2)$ with $P_{\text{mean}} = -1.2$ and $P_{\text{std}} = 1.2$. The network is conditioned on $c_{\text{noise}} = \tfrac{1}{4}\ln\sigma$. At sampling time we use the discretization with $\sigma_{\min} = 0.002$, $\sigma_{\max} = 80$, and $\rho = 7$, integrated with a first-order Euler solver. The deterministic variant (ODE) integrates the probability-flow ODE, and the stochastic variant (SDE) adds stochastic churn with $S_{\text{churn}} = 10$, $S_{\text{noise}} = 1$, $S_{t\min} = 0$, and $S_{t\max} = 10^6$.

\paragraph{Flow Matching.} We use the linear Gaussian path $\mathbf{y}_s = (1-s)\,\mathbf{x}_1 + s\,\boldsymbol{\epsilon}$, with continuous $s \sim \mathcal{U}(10^{-5}, 1]$. The network regresses the velocity $\boldsymbol{\epsilon} - \mathbf{x}_1$ with an MSE loss. Sampling integrates from $s = 1$ to $s = 0$ with a uniform-step Euler solver.

\paragraph{Stochastic Interpolants.} We use the data-to-data interpolant
\begin{equation*}
    \mathbf{I}_s = (1-s)\,\mathbf{x}_0 + s\,\mathbf{x}_1 + \sigma(s)\,\mathbf{W}_s, \qquad \sigma(s) = \sigma_0 (1-s), \qquad \mathbf{W}_s = \sqrt{s}\,\boldsymbol{\epsilon},
\end{equation*}
with $s \sim \mathcal{U}[0, 1)$. The network regresses the drift $\partial_s \mathbf{I}_s = \mathbf{x}_1 - \mathbf{x}_0 + \dot{\sigma}(s)\,\mathbf{W}_s$. We use antithetic sampling ($\pm\mathbf{W}_s$) to reduce the variance of the loss \citep{Albergo2023}. Sampling starts at the conditioning state $\mathbf{x}_0$. The first step is taken analytically, and the remaining steps are integrated from $s = 0$ to $s = 1$ with either Euler (ODE, no injected noise) or Euler--Maruyama with diffusion coefficient $\sigma(s)$ (SDE). The noise scale is $\sigma_0 = 1.0$ for Kolmogorov Flow and $\sigma_0 = 0.5$ for Rayleigh-B\'enard and PlaSim.

\subsection{Architectures}
\label{app:arch}
All generative models run in the latent space of a pretrained autoencoder. The autoencoder consists of a CNN-based encoder and decoder, which rely on ResNet-style blocks and pooling/unpooling layers to compress and reconstruct the physical states. Residual pooling/unpooling is used, following \citet{chen2025deepcompressionautoencoderefficient}. The latent space has additional regularization to prevent arbitrary variance. 

After pretraining, the autoencoder is frozen, and the generative model learns the one-step map $\mathbf{z}_t \mapsto \mathbf{z}_{t+\Delta t}$ between consecutive latent states. All models share the same backbone, which is a Diffusion Transformer (DiT). The diffusion or interpolant time $t$ is scaled by $10^3$, given a sinusoidal embedding, and injected through adaptive layer norm (adaLN). Physical parameters, when present, are embedded the same way and added to the time embedding. The model input is the channel-wise concatenation of the conditioning latent $\mathbf{z}_t$ and the current noisy or interpolated state, so the input width is twice the latent width. 

Transformer backbones are trained with Muon on the hidden weight matrices of the transformer body, at $10\times$ the base learning rate. All remaining parameters (embeddings, adaLN modulators, output head, and 1D gains and biases) use AdamW at the base learning rate. We keep an exponential moving average (EMA) of the weights, and all validation and evaluation use the EMA weights. Models are trained to the same gradient budget within each experiment; this is 25,000/60,000/50,000 steps for Kolmogorov Flow, Rayleigh B\'enard, and PlaSim, respectively. Lastly, the main hyperparameters for the architectures used in each set of experiments are given in Table \ref{tab:hyperparameters}. 

\begin{table}[h]
    \centering
    \begin{tabular}{llll}
        \toprule
        Parameter & Kolmogorov Flow & Rayleigh-Benard & PlaSim \\
        \midrule
        \multicolumn{4}{l}{\textit{Autoencoder}} \\
        Input resolution & $160 \times 160 \times 1$ & $512 \times 128 \times 4$ & $64 \times 128 \times 73$ \\
        Latent resolution & $20 \times 20 \times 4$ & $64 \times 16 \times 4$ & $16 \times 32 \times 32$ \\
        Downsampling factor & 64 & 64 & 32 \\
        Channel widths & [64, 128, 256, 256] & [64, 128, 256, 512] & [128, 256, 512] \\
        ResBlocks per stage & 3 & 4 & 2 \\
        Parameters & 20.5M & 57.3M & 61.1M \\
        \midrule
        \multicolumn{4}{l}{\textit{DiT}} \\
        Hidden dim & 512 & 1024 & 1024 \\
        Heads & 16 & 16 & 16 \\
        Layers & 12 & 12 & 12 \\
        Patch size & $1 \times 1$ & $2 \times 2$ & $1 \times 1$ \\
        Tokens & 400 & 256 & 512 \\
        Parameters & 57.9M & 232.7M & 313.1M \\
        \bottomrule
    \end{tabular}
    \caption{DiT and autoencoder hyperparameters across all experiments.}
    \label{tab:hyperparameters}
\end{table}

\section{Additional Results}

\subsection{Performance vs. NFEs}
\label{app:nfes}
\begin{figure}
    \centering
    \includegraphics[width=\linewidth]{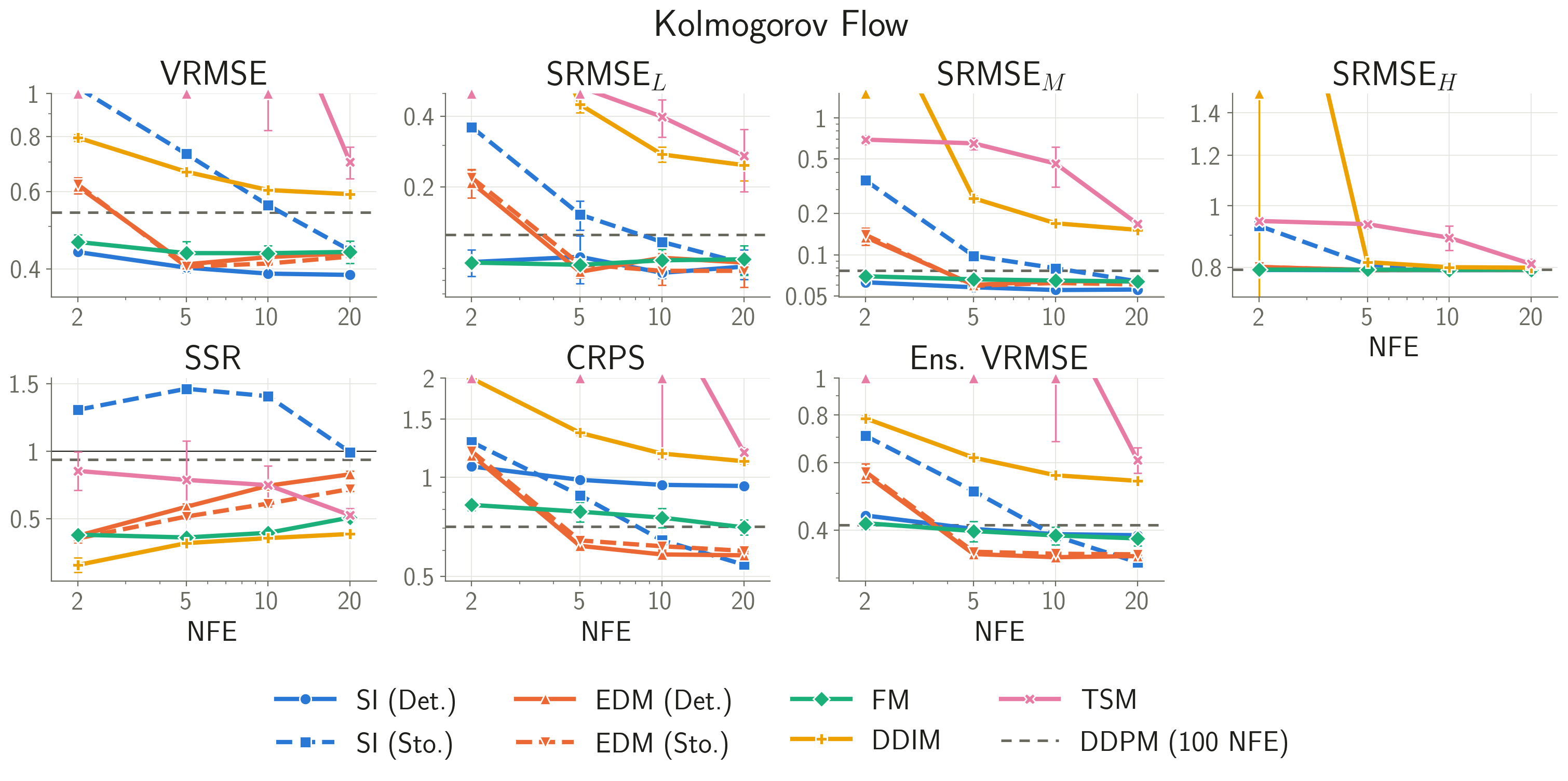}
    \caption{Model performance vs. NFEs for Kolmogorov Flow, plotted for different metrics. The mean over three seeds is plotted, with the error bars showing the standard deviation. Maximum values on the y-axes are thresholded and the performance of DDPM is denoted by a horizontal line. }
    \label{fig:km_flow_nfe}
\end{figure}

\begin{figure}
    \centering
    \includegraphics[width=\linewidth]{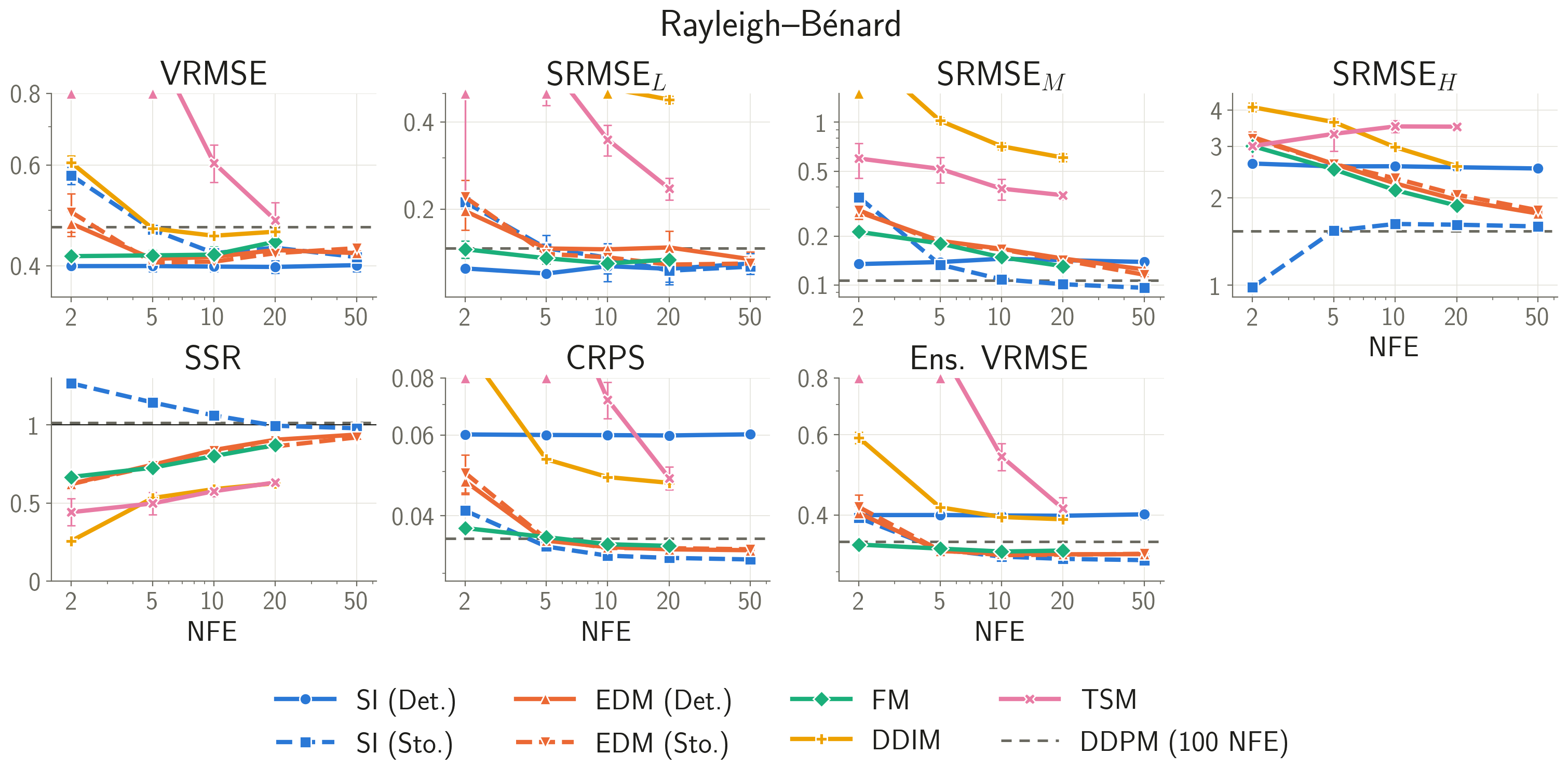}
    \caption{Model performance vs. NFEs for Rayleigh-B\'enard, plotted for different metrics. The mean over three seeds is plotted, with the error bars showing the standard deviation. Maximum values on the y-axes are thresholded and the performance of DDPM is denoted by a horizontal line. }
    \label{fig:rb_nfe}
\end{figure}

\begin{figure}
    \centering
    \includegraphics[width=\linewidth]{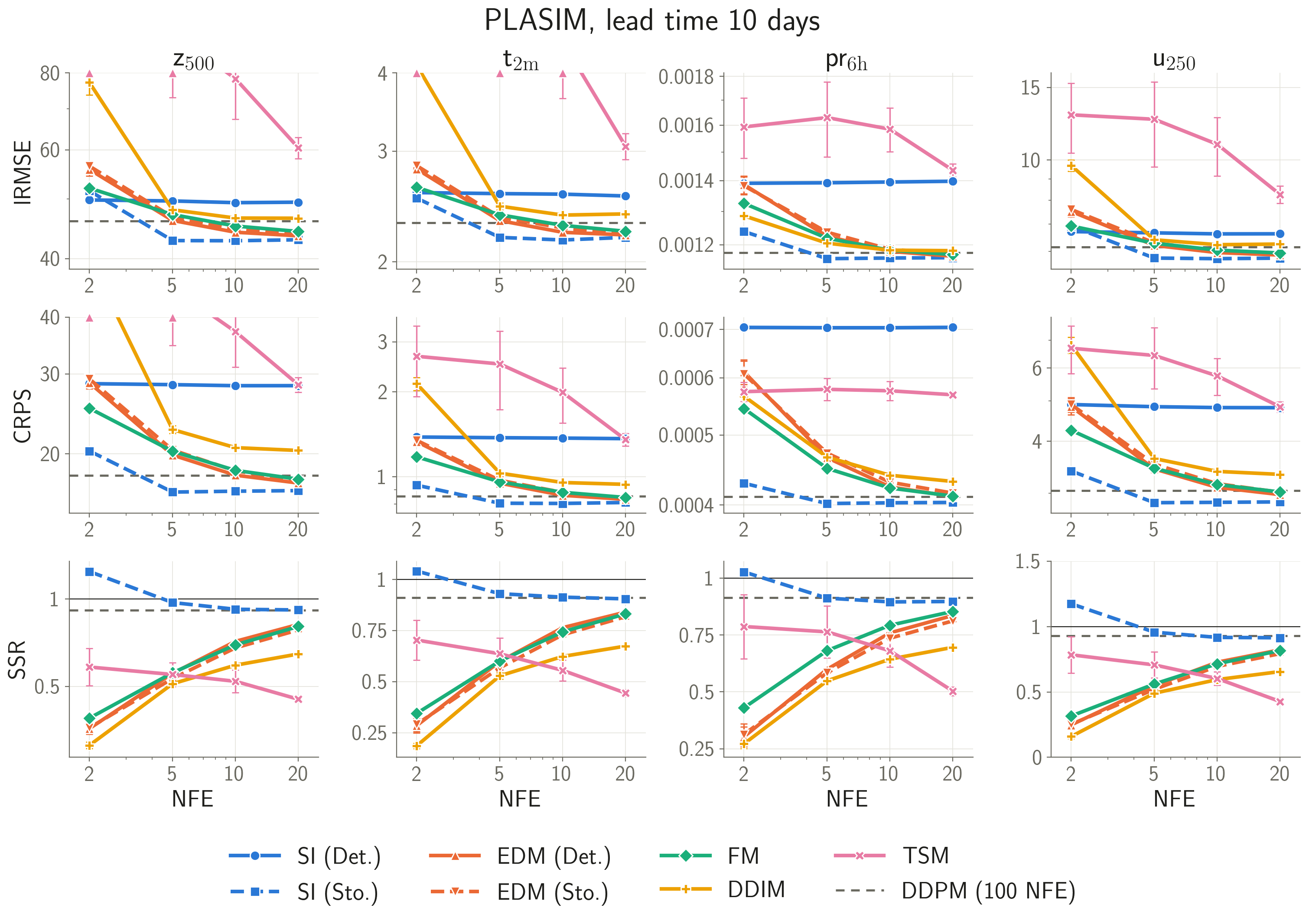}
    \caption{Model performance vs. NFEs for PlaSim plotted for different metrics. The metrics are evaluated at a 10-day lead time, and the lRMSE is computed using the ensemble mean. The mean over three seeds is plotted, with the error bars showing the standard deviation. Maximum values on the y-axes are thresholded and the performance of DDPM is denoted by a horizontal line. }
    \label{fig:plasim_nfe}
\end{figure}

The main paper reports results at a fixed NFE for clarity; however, in this section, we perform a sweep over \(\text{NFE}\in\{2,5,10,20\}\) for each model and seed, and compute the relevant metrics. These results are shown in Figure \ref{fig:km_flow_nfe} for Kolmogorov Flow, Figure \ref{fig:rb_nfe} for Rayleigh B\'enard, and Figure \ref{fig:plasim_nfe} for PlaSim. PlaSim metrics are computed at a 10-day lead time. All plots visualize the mean over three seeds, with error bars denoting the standard deviation. 

In general, we can see that SI models tend to perform well, although they are not the best at every metric and at every NFE. For example, on Kolmogorov Flow and Rayleigh-B\'enard, the stochastic sampler (SI (Sto.)) can need more sampling steps to achieve good probabilistic performance (low CRPS and ensemble VRMSE, SSR \(\approx\! 1\)). Models such as flow matching or EDM occasionally outperform SI models as well. However, on the PlaSim benchmark, we can see that SI models consistently achieve the best performance across the evaluated NFEs. Also, we can see that SI models are the only framework that is capable of being over-dispersive (SSR\(>1\)), which is consistent across NFEs. 

As a whole, we also observe that most metrics improve with increasing sampling steps and plateau at a certain point. Performance at low NFEs (\(\sim\!2,5\)) can also be quite poor for many generative models. Finally, the spread between seeds is quite small, especially compared to the effect of increasing NFEs or switching to different generative frameworks. 

\subsection{Wall Clock Time}
\label{app:wall_clock}
\begin{figure}
    \centering
    \includegraphics[width=\linewidth]{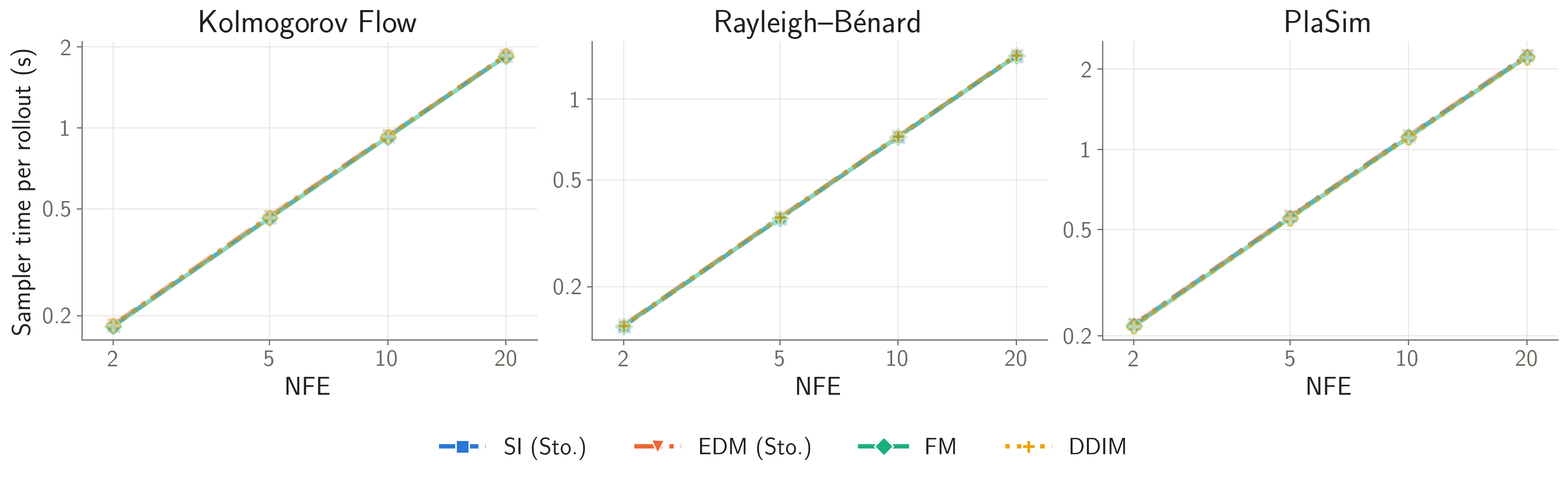}
    \caption{Wall clock time vs. number of function evaluations (NFEs), for different generative models and datasets. Since each generative model uses the same architecture, they take the same amount of time at a given NFE. The time to generate a rollout also scales linearly with the NFEs.}
    \label{fig:wallclock}
\end{figure}

In the main results, we report the performance of models as a function of the number of sampling steps, however, this may not correspond to the actual wall-clock time. Therefore, in this section, we run a set of experiments where we measure the wall-clock time of generating a single rollout, as a function of NFEs. This is plotted in Figure \ref{fig:wallclock}, where the wall clock time is measured on a single H100 GPU. In general, we see that each model takes the same amount of time to generate a rollout, for a given NFE and experiment. This is because each model is uses the same architecture and model size within an experiment. Furthermore, the NFEs track the wall clock time closely, where they are correlated linearly. During generation, there are no dataloading bottlenecks, so increasing NFEs directly corresponds to an increase in the compute time. 

\subsection{Performance vs. Prediction stride \(\Delta t\)}
\label{app:stride}

We present an additional set of results where we vary the prediction stride \(\Delta t\) of generative models on Kolmogorov Flow. Given a full rollout up to \(t=20\) with 100 timesteps, we consider \(\Delta t = \{0.2, 0.4,0.8s\}\) to see the effect on rollout errors and the relative ranking of generative models. The results are given in Table \ref{tab:km_flow_stride}. We can see the increasing the stride slightly improves model performance, potentially due to accumulating less autoregressive error. However, this usually dataset-dependent and is a tradeoff: as \(\Delta t\) increases the next state is more difficult to predict, but there is less autoregressive error accumulation \citep{aceituno2026temporalhorizonsforecastingperformancelearnability, couairon2026kastorefficientfinetuningstrategy}. In our case, the stride \(\Delta t\) is also important for the SI model, as states can start to be further apart as the stride increases. This can reduce the advantage of coupled source and target distributions, and we see that EDM and other models start to close the gap to SI models at larger \(\Delta t\), though this effect is small. 

\begin{table}[t]
    \centering
    \resizebox{\linewidth}{!}{%
    \begin{tabular}{l l | c c c c | c c c}
        \toprule
        Model & NFE & VRMSE & SRMSE$_L$ & SRMSE$_M$ & SRMSE$_H$ & SSR & CRPS & Ens.\ VRMSE \\
        \midrule
        \multicolumn{9}{l}{\textit{$\Delta t = 0.2$ (100-step rollout)}} \\
        \midrule
        FNO             & 1   & 0.591 & \cellcolor{mygray}0.070 & \cellcolor{mygray}0.043 & \cellcolor{mygray}0.398 & --- & 1.522 & 0.591 \\
        DDPM            & 100 & 0.539 & 0.122 & 0.079 & 0.795 & \cellcolor{mygray2}0.952 & 0.702 & 0.411 \\
        DDIM            & 10  & 0.605 & 0.260 & 0.171 & 0.802 & 0.363 & 1.172 & 0.554 \\
        TSM             & 10  & 1.004 & 0.411 & 0.336 & 0.854 & 0.591 & 1.603 & 0.845 \\
        EDM (Det.)      & 10  & 0.423 & 0.099 & 0.065 & \cellcolor{mygray2}0.793 & 0.771 & \cellcolor{mygray2}0.574 & \cellcolor{mygray2}0.337 \\
        EDM (Sto.)      & 20  & 0.427 & \cellcolor{mygray2}0.078 & 0.063 & 0.794 & 0.738 & 0.601 & 0.349 \\
        FM              & 10  & \cellcolor{mygray2}0.415 & 0.092 & 0.062 & \cellcolor{mygray2}0.793 & 0.419 & 0.697 & 0.365 \\
        SI (Det.) & 10  & \cellcolor{mygray}0.383 & 0.084 & \cellcolor{mygray2}0.056 & \cellcolor{mygray2}0.793 & --- & 0.929 & 0.383 \\
        SI (Sto.) & 20  & 0.431 & 0.110 & 0.063 & \cellcolor{mygray2}0.793 & \cellcolor{mygray}1.000 & \cellcolor{mygray}0.536 & \cellcolor{mygray}0.326 \\
        \midrule
        \multicolumn{9}{l}{\textit{$\Delta t = 0.4$ (50-step rollout)}} \\
        \midrule
        FNO             & 1   & 0.567 & \cellcolor{mygray}0.083 & \cellcolor{mygray}0.042 & \cellcolor{mygray}0.410 & --- & 1.441 & 0.567 \\
        DDPM            & 100 & 0.540 & 0.111 & 0.080 & 0.795 & 1.111 & 0.643 & 0.386 \\
        DDIM            & 10  & 0.578 & 0.211 & 0.145 & 0.799 & 0.458 & 1.022 & 0.511 \\
        TSM             & 10  & 0.623 & 0.187 & 0.131 & 0.801 & 0.669 & 0.969 & 0.519 \\
        EDM (Det.)      & 10  & 0.416 & 0.092 & 0.064 & 0.794 & \cellcolor{mygray}0.933 & \cellcolor{mygray}0.488 & \cellcolor{mygray}0.301 \\
        EDM (Sto.)      & 20  & 0.406 & 0.096 & 0.058 & 0.793 & 0.913 & 0.502 & \cellcolor{mygray2}0.307 \\
        FM              & 10  & \cellcolor{mygray2}0.401 & 0.097 & 0.061 & 0.794 & 0.606 & 0.607 & 0.346 \\
        SI (Det.) & 10  & \cellcolor{mygray}0.369 & \cellcolor{mygray2}0.085 & \cellcolor{mygray2}0.052 & \cellcolor{mygray2}0.792 & --- & 0.887 & 0.369 \\
        SI (Sto.) & 20  & 0.420 & 0.087 & 0.059 & 0.793 & \cellcolor{mygray2}1.084 & \cellcolor{mygray2}0.500 & 0.310 \\
        \midrule
        \multicolumn{9}{l}{\textit{$\Delta t = 0.8$ (25-step rollout)}} \\
        \midrule
        FNO             & 1   & 0.572 & \cellcolor{mygray}0.061 & \cellcolor{mygray}0.044 & \cellcolor{mygray}0.438 & --- & 1.464 & 0.572 \\
        DDPM            & 100 & 0.497 & 0.113 & 0.071 & 0.794 & 1.207 & 0.609 & 0.368 \\
        DDIM            & 10  & 0.505 & 0.163 & 0.119 & 0.798 & 0.665 & 0.752 & 0.416 \\
        TSM             & 10  & 0.581 & 0.123 & 0.082 & 0.796 & 0.783 & 0.803 & 0.451 \\
        EDM (Det.)      & 10  & \cellcolor{mygray2}0.396 & 0.069 & 0.057 & 0.793 & \cellcolor{mygray2}1.035 & \cellcolor{mygray2}0.484 & \cellcolor{mygray}0.300 \\
        EDM (Sto.)      & 20  & 0.401 & 0.083 & 0.063 & 0.794 & \cellcolor{mygray}1.029 & 0.486 & \cellcolor{mygray}0.300 \\
        FM              & 10  & 0.421 & 0.101 & 0.067 & 0.794 & 0.883 & 0.540 & 0.323 \\
        SI (Det.) & 10  & \cellcolor{mygray}0.353 & \cellcolor{mygray2}0.063 & \cellcolor{mygray2}0.051 & \cellcolor{mygray2}0.792 & --- & 0.835 & 0.353 \\
        SI (Sto.) & 20  & 0.415 & 0.086 & 0.061 & 0.793 & 1.128 & \cellcolor{mygray}0.479 & \cellcolor{mygray2}0.301 \\
        \bottomrule
    \end{tabular}}
    \caption{Effect of the dataset stride on Kolmogorov Flow. Each model is trained at
    $\Delta t \in \{0.2, 0.4, 0.8\}$ and rolled out over the same physical horizon
    ($t_{final} = 20$). 
    $L$, $M$, $H$ denote low-, mid-, and high-band SRMSE.
    VRMSE and SRMSE are computed per ensemble member; SSR, CRPS, and Ens.\ VRMSE (VRMSE of the ensemble mean) use 8-member ensembles.
    SSR is undefined (---) for deterministic models, whose Ens.\ VRMSE is that of their single forecast.
    Shading marks the best and second-best entry per column within each $\Delta t$ block. Results are for a single seed.}
    \label{tab:km_flow_stride}
\end{table}

\subsection{Transport Cost}
\label{app:transport}

\paragraph{Overview} A hypothesis of this work is that the distributions of current and future states are closer together than Gaussian noise and future states, which allows stochastic interpolants to be more efficient or accurate than conditional generation. Consider an example where initial states are uniformly sampled from a set of initial conditions. For dissipative PDEs, this uniform distribution is transported over time to a stationary distribution as energy is lost. If the time interval is small, we may intuitively believe that subsequent distributions are close; however, we seek to visualize and loosely quantify this difference. 

Calculating statistical distances between high-dimensional empirical distributions is usually intractable. Despite this, there are several heuristics that are used. For example, Fréchet Inception Distance (FID) \citep{heusel2018ganstrainedtimescaleupdate} computes distances based on activations of a neural network; while this has no mathematical basis, it is useful to estimate distances between image distributions. We consider more general heuristics such as the Sliced Wasserstein Distance (SW), Classifier 2-Sample Test (C2ST), and Maximum Mean Discrepancy (MMD) \citep{bischoff2024practicalguidesamplebasedstatistical}. For example, C2ST trains a classifier to discriminate samples drawn from two distributions; if the classifier is perfect (\(100\%\) accuracy), then the distributions can be viewed as farther than if the classifier cannot identify samples (\(50\%\) accuracy). 

\begin{figure}[t!]
    \centering
    \includegraphics[width=\linewidth]{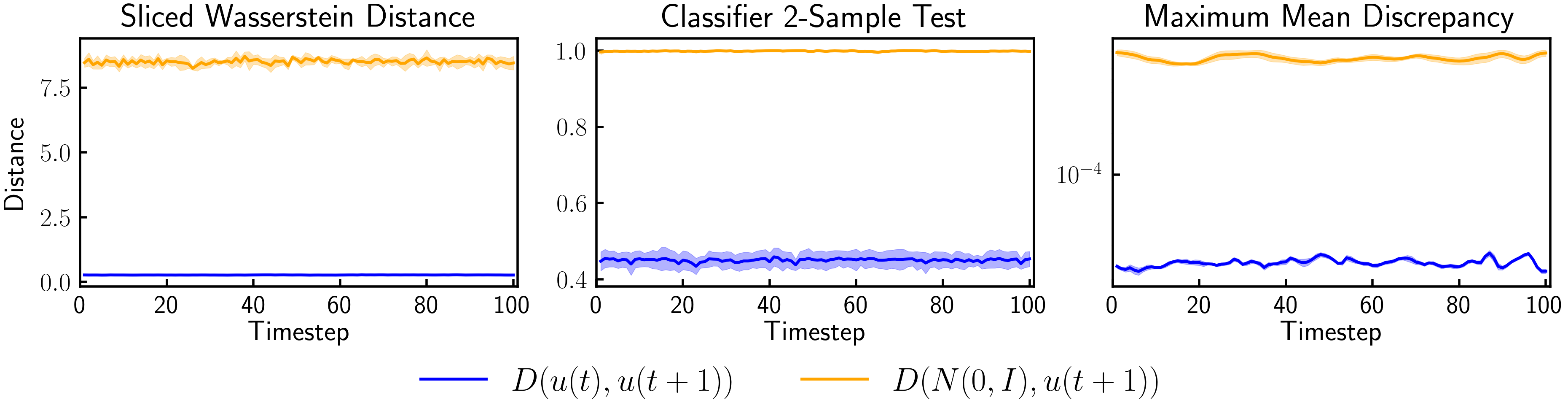}
    \caption{Distance heuristics for the Kolmogorov Flow dataset. Distances between samples drawn from successive timesteps \(D(u(t), u(t+1))\) and between Gaussian noise and future timesteps \(D(N(0, I), u(t+1))\) are plotted for the vorticity field. Each metric is averaged over a 5-fold cross validation, with the standard deviation shaded.}
    \label{fig:distances_kf}
\end{figure}

\begin{figure}[t!]
    \centering
    \includegraphics[width=\linewidth]{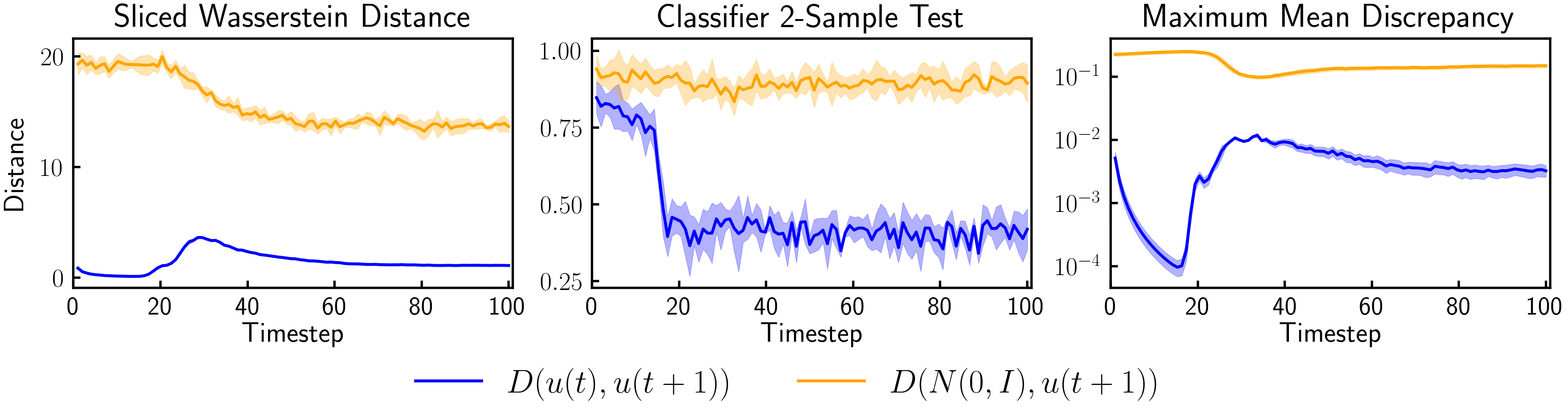}
    \caption{Distance heuristics for the Rayleigh-Bénard dataset at \(Ra=1e10\) and \(Pr=0.1\). Distances between samples drawn from successive timesteps \(D(u(t), u(t+1))\) and between Gaussian noise and future timesteps \(D(N(0, I), u(t+1))\) are plotted for the buoyancy field. Each metric is averaged over a 5-fold cross validation, with the standard deviation shaded.}
    \label{fig:distances}
\end{figure}

We adopt a perspective where each timestep \(\mathbf{u}_t \in \mathbb{R}^{n_x \times n_y}\) of a dataset is sampled from a different underlying distribution \(\mathbf{u}_t\sim \rho _t\). While \(\rho_t\) is unknown, we have access to an empirical distribution with \(n\) samples \(\rho^n_t = \{\mathbf{u}^{1}_t, \ldots \mathbf{u}^{n}_t\}\), where \(n\) is the dataset size. Heuristics are computed using these empirical distributions as well as \(n\) samples \(\mathbf{z}\in \mathbb{R}^{n_x \times n_y}\) drawn from a Gaussian \(\mathbf{z} \sim \mathcal{N}(0, \mathbf{I})\). For the Kolmogorov Flow and Rayleigh Benard dataset, we calculate each heuristic \(D\) over time for either \(D(\rho^n_t, \rho^n_{t+1})\) or \(D(\mathcal{N}(0, \mathbf{I}), \rho^n_{t+1})\). Results are plotted and shown in Figures \ref{fig:distances_kf} and \ref{fig:distances}

In general, the heuristics suggest that distances between subsequent timesteps are closer than distances between future timesteps and Gaussian noise. This is further corroborated by physical intuition, where Kolmogorov flows are known to reach a stationary distribution \citep{PhysRevE_km_flow} due to its periodic boundary conditions and forcing. Therefore, in these scenarios, it may be beneficial to train interpolants rather than generative models based on Gaussian denoising. However, there are many confounding factors. Firstly, distributions that are far by a certain metric (KL-Divergence, FID, C2ST, etc.) does not mean that their transport map is difficult to learn, such as two Gaussians offset by an arbitrarily large mean. Furthermore, modern deep learning has made learning complex transport maps feasible, to the point that differences between a Gaussian and a well-designed source distribution may not matter in practice. 

\end{document}